\documentclass[10pt,twocolumn,letterpaper]{article}

\usepackage[pagenumbers]{wacv} 

\usepackage{graphicx}
\usepackage{amsmath}
\usepackage{amssymb}
\usepackage{booktabs}
\usepackage{algorithm}
\usepackage{algpseudocode}
\usepackage{xcolor}
\usepackage{mathtools}
\usepackage{graphicx}
\usepackage{transparent}

\usepackage[pagebackref,breaklinks,colorlinks]{hyperref}

\usepackage[capitalize]{cleveref}
\crefname{section}{Sec.}{Secs.}
\Crefname{section}{Section}{Sections}
\Crefname{table}{Table}{Tables}
\crefname{table}{Tab.}{Tabs.}

\def\wacvPaperID{536} 
\def\confName{WACV}
\def\confYear{2024}

\newcommand{\bx}[0]{\mathbf{x}}
\newcommand{\bxz}[0]{\mathbf{x}_0}
\newcommand{\bxzhat}[0]{\hat{\mathbf{x}}_0}
\newcommand{\bxt}[0]{\mathbf{x}_t}
\newcommand{\bxtt}[1]{\mathbf{x}_{#1}}

\newcommand{\bI}[0]{\mathbf{I}}
\newcommand{\by}[0]{\mathbf{y}}

\newcommand{\atbar}[0]{\bar{\alpha}_t}

\newcommand{\sqrtoatbar}[0]{\sqrt{1-\atbar}}

\begin{document}

\hypersetup{
	colorlinks=true,
	linkcolor=blue,
	filecolor=blue,
	citecolor=green,
	urlcolor=blue,
	breaklinks=false
}

\title{Nested Diffusion Processes for Anytime Image Generation}

\author{Noam Elata$^{1}$ \quad Bahjat Kawar$^{2}$ \quad Tomer Michaeli$^{1}$ \quad Michael Elad$^{2}$\\
$^{1}$Department of ECE \quad $^{2}$Department of CS \\
Technion -- Israel Institute of Technology \\
{\tt\small \{noamelata@campus, bahjat.kawar@cs, tomer.m@ee, elad@cs\}.technion.ac.il}
}

\maketitle

\begin{abstract}
Diffusion models are the current state-of-the-art in image generation, synthesizing high-quality images by breaking down the generation process into many fine-grained denoising steps. 
Despite their good performance, diffusion models are computationally expensive, requiring many neural function evaluations (NFEs). 
In this work, we propose an anytime diffusion-based method that can generate viable images when stopped at arbitrary times before completion.
Using existing pretrained diffusion models, we show that the generation scheme can be recomposed as two nested diffusion processes, enabling fast iterative refinement of a generated image. 
In experiments on ImageNet and Stable Diffusion-based text-to-image generation, we show, both qualitatively and quantitatively, that our method's intermediate generation quality greatly exceeds that of the original diffusion model, while the final generation result remains comparable. 
We illustrate the applicability of Nested Diffusion in several settings, including for solving inverse problems, and for rapid text-based content creation by allowing user intervention throughout the sampling process.\footnotemark

\end{abstract}

\section{Introduction}

Diffusion models (DMs) have emerged as a promising class of generative models ~\cite{sohl2015deep, ho2020denoising, song2019generative}. Having achieved state-of-the-art capabilities in image generation, DMs have also excelled at various tasks, such as inverse problem solving~\cite{kawar2021stochastic, kawar2022denoising, lugmayr2022repaint, wang2022zero, chung2023diffusion, song2023pseudoinverse}, and image editing~\cite{meng2021sdedit, hertz2022prompt, brooks2022instructpix2pix, kawar2023imagic}. 
DMs have also been successfully applied to a wide range of domains, ranging from speech and audio~\cite{kong2020diffwave, jeong2021diff} to protein structures~\cite{qiao2022dynamic, schneuing2022structure} and medical data~\cite{song2023solving, chung2022score, kawar2023gsure}.

The sampling process of modern DMs can be computationally expensive~\cite{salimans2022progressive, song2020denoising, lu2022dpmpp}, due to the large networks used and the iterative nature of the reverse diffusion process. Despite the progress in acceleration of DMs~\cite{song2020denoising, salimans2022progressive, song2023consistency}, many of the leading diffusion model-based applications remain prohibitively slow.

During sampling, the diffusion process creates intermediate image predictions as a byproduct, denoted as $\bxzhat$, at various time steps.  In theory, this allows for monitoring the generation process and assessing the resulting images without waiting for its completion. However, these predictions do not align with the learned image manifold and often exhibit a smooth or blurry appearance~\cite{kawar2021stochastic}.

To address this issue, we propose Nested Diffusion, a novel technique that leverages a pretrained DM to iteratively refine generated images, acting as an \emph{anytime} generation algorithm. With Nested Diffusion, intermediate predictions of the final generated image are of better quality, which allows users to observe the generated image during the sampling process and to conclude the generation if the intermediate yielded image is already satisfactory. Through experiments, we observe that our Nested Diffusion shows superior intermediate generation quality compared to the original DM, while maintaining comparable final results.

Access to high quality intermediate predictions is advantageous in several DM-based applications beyond image generation. 
Nested Diffusion's anytime algorithm is also beneficial in solving inverse problems, given the prominence of DMs in this area.
In scenarios where multiple output images are generated, users can control the generation process by selecting the most promising candidate and influencing the sampling process to prioritize their preferred images. 
This valuable capability also enables interactive online adjustments during the generation process.

We introduce Nested Diffusion, a novel anytime sampling algorithm for DMs. We showcase the versatility of our method through various applications, including conditional image generation, and inverse problem solving. 
Additionally, we highlight the ability of Nested Diffusion to enable interactive content creation during the sampling process. \footnotetext{Code available at \urlstyle{sf}\url{https://github.com/noamelata/NestedDiffusion}.}

\begin{figure*}[t]
\centering
    \includegraphics[width=6.8in]{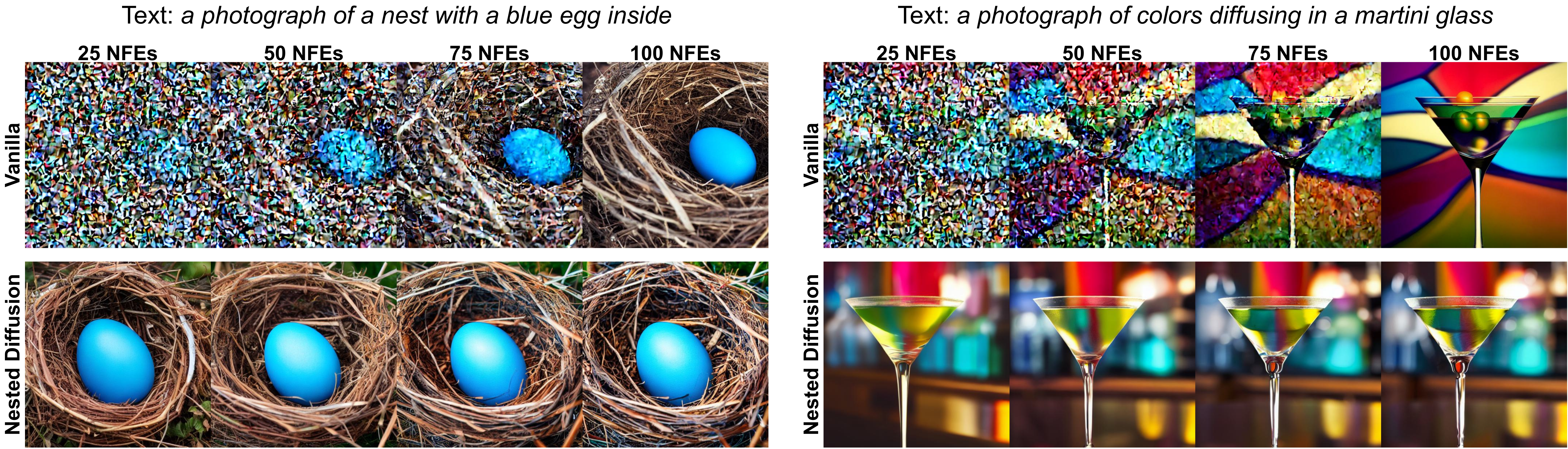}
    \vskip 0.2in
    \includegraphics[width=6.8in]{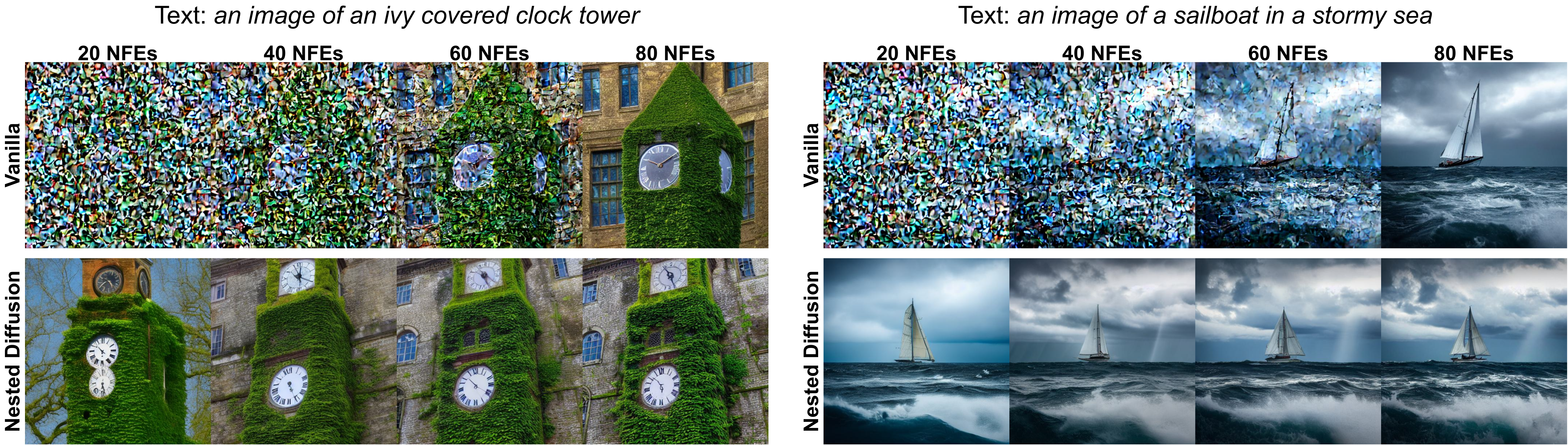}
\caption{\textbf{Results of intermediate predictions of Stable Diffusion from a diffusion process of 100 NFEs (top) and 80 NFEs (bottom).}}
\label{fig:stable}

\vskip -0.1in
\end{figure*}

\section{Preliminaries}

\subsection{Anytime Algorithms}
Anytime algorithms~\cite{boddy1989solving, zilberstein1996using, grass1996anytime, horsch2013anytime} are a class of methods that attempt to address real-world problems under resource constraints, time limitations, or uncertain input information. 
Specifically, these algorithms generate progressively improved solutions, which enable early interruption. 
Unlike conventional algorithms that require completion for a final solution, anytime algorithms offer users the flexibility to obtain usable solutions at any stage of execution. 
This adaptability proves highly valuable in time-sensitive decision-making and iterative problem-solving scenarios. 

\subsection{Diffusion Models}
Diffusion models (DMs)~\cite{sohl2015deep, ho2020denoising, song2019generative} are the state-of-the-art generative models~\cite{dhariwal2021diffusion}, relying on the capabilities of deep neural networks (DNN) in removing Gaussian noise.
The forward diffusion process is defined as a degradation of a data point $\bxz$ in a dataset $\mathcal{D}$ with accumulated Gaussian noise through the process $q(\bxt|\bxtt{t-1})=\mathcal{N}(\bxt; \sqrt{1-\beta_t}\bxtt{t-1},\beta_t \bI)$ defined over timesteps $t=1, \ldots, T$, where $\beta_t$ are the noise amplitudes. 
During training, the reverse diffusion process $p_\theta(\bxtt{t-1}|\bxt)$ is learned by maximizing the evidence lower bound (ELBO) on the training dataset. The ELBO can be written in terms of Kullback Leibler divergence terms between $q(\bxtt{t-1} | \bxz, \bxt)$ and $p_\theta(\bxtt{t-1}|\bxt)$, which have a simple closed-form expression when $p_\theta(\bxtt{t-1}|\bxt)$ is modeled as a  Gaussian distribution.
At inference time, the trained DNN gradually removes noise from a random initialization $\bxtt{T}\sim\mathcal{N}(0, \bI)$, sampling iteratively from the learned distributions $p_\theta(\bxtt{t-1}|\bxt)$, and finally outputting a generated image from a distributions approximating the real image distribution, $\bxz \sim q(\bxz)$.

\section{Nested Diffusion}
\subsection{Formulation}

In DDPM~\cite{ho2020denoising},  $p_\theta(\bxtt{t-1}|\bxt)$ is assumed to follow a Gaussian distribution, with its mean defined using the expectation $\mathbb{E}[\bxz | \bxt]$ yielded by the DNN, and its variance defined as a constant. Thus, we can sample from $p_\theta(\bxtt{t-1}|\bxt)$ in closed-form.
However, we can also interpret this sampling process by expressing the distribution $p_\theta(\bxtt{t-1}|\bxt)$ as a convolution of two others~\cite{xiao2021tackling} -- the distribution $q(\bxtt{t-1}|\bxz, \bxt)$ which has a closed form, and the DNN-based approximation  $p_\theta(\bxz|\bxt)$, 
\begin{equation}
    \label{eq:marginalization_p_x_t_1_x_t}
    p_\theta(\bxtt{t-1}|\bxt) = \int q(\bxtt{t-1} | \bxz, \bxt) p_\theta(\bxz | \bxt) d\bxz.
\end{equation}
Here, $p_\theta(\bxz | \bxt)$ is the distribution represented by the DNN in the context of predicting $\bxz$. For instance, DMs with a deterministic DNN output, such as DDPM, would correspond to a Dirac delta function distribution around the DNN-estimated $\mathbb{E}[\bxz | \bxt]$. The stochasticity in the reverse diffusion process would be obtained by setting  $q(\bxtt{t-1} | \bxz, \bxt)$ as a fixed Gaussian.
More generally, sampling from the joint distribution ${p_\theta(\bxtt{t-1}, \bxz|\bxt) = q(\bxtt{t-1} | \bxz, \bxt) p_\theta(\bxz | \bxt)}$
can be done sequentially, by first sampling $\bxzhat \sim p_\theta(\bxz | \bxt)$ and then sampling $\bxtt{t-1}\sim q(\bxtt{t-1} | \bxzhat, \bxt)$, yielding $\bxtt{t-1}$ that follows \Cref{eq:marginalization_p_x_t_1_x_t}. The generalized reverse diffusion process, following this interpretation, is presented in \Cref{alg:diffusion-sampling}.

\begin{algorithm}[t]
\centering
\caption{Sampling from a Regular Diffusion Process}
\label{alg:diffusion-sampling}
\begin{algorithmic}
\State $\bxtt{T} \sim \mathcal{N}(0, \bI)$
\For {$t$ in $\{T, T-s, \ldots, 1+s,  1\}$}
\State $\bxzhat \sim p_\theta(\bxz | \bxt) $
\State $\bxtt{t-s} \sim q(\bxtt{t-s} | \bxzhat, \bxt)$
\EndFor
\State return $\bxz$
\end{algorithmic}
\end{algorithm}

Note that after training $p_\theta(\bxz|\bxt)$ for a certain $q(\bxtt{t-1} | \bxz, \bxt)$, it is possible to utilize the same DNN model for different distributions $q$.
For instance, DDIM~\cite{song2020denoising} utilizes a deterministic $q(\bxtt{t-1} | \bxz, \bxt)$ (equivalent to a Dirac delta function) for faster generation.
Interestingly, by sampling using \Cref{alg:diffusion-sampling}, the Gaussian assumption on $p_\theta(\bxtt{t-1}|\bxt)$ is no longer required, and can be generalized beyond DDPM sampling. 
In this setting, $p_\theta(\bxz|\bxt)$ may be any learned distribution, and is not restricted to a delta function or a Gaussian form.

\subsection{Method}
\label{seq:method}
We suggest that many valid choices of $q(\bxtt{t-1} | \bxz, \bxt)$ and an accurate DNN-based approximation $p_\theta(\bxz|\bxt)$ 
can generate high quality samples using ~\Cref{eq:marginalization_p_x_t_1_x_t} and ~\Cref{alg:diffusion-sampling}. 
This could allow us to harness many different generative models into the diffusion process, for instance as done with GANs~\cite{goodfellow2014generative} by  Xiao et al.~(2022)~\cite{xiao2021tackling}. Here, we we propose to use a complete diffusion process as a good approximation for $p_\theta(\bxz|\bxt)$.

\begin{algorithm}[t]
\centering
\caption{Sampling from Nested Diffusion}
\label{alg:nested-diffusion-sampling}
\begin{algorithmic}
\State Outer diffusion denoted in \textcolor{blue}{blue}, with step size $s^o$
\State Inner diffusion denoted in \textcolor{purple}{purple}, with step sizes $\{s^i_t\}$
\State $\mathcolor{blue}{\bxtt{T}} \sim \mathcal{N}(0, \bI)$
\For{$t$ in $\{T, T-s^o,  \ldots, 1+s^o, 1\}$}
\State $\mathcolor{purple}{\bxt'} = \mathcolor{blue}{\bxt}$
\Comment{\textcolor{gray}{Beginning of inner diffusion}}
\For{$\tau$ in $\{t, t-s^i_t,  \ldots, 1+s^i_t, 1\}$}
\State $\mathcolor{purple}{\hat{\bx}_{0}'} \sim \mathcolor{purple}{p_\theta(\bxtt{0}' | \bxtt{\tau}')}$
\State $\mathcolor{purple}{\bxtt{\tau - s^i_t}'} \sim \mathcolor{purple}{q'(\bxtt{\tau - s^i_t}' | \hat{\bx}_{0}', \bxtt{\tau}')}$
\EndFor
\State $\mathcolor{blue}{\bxzhat} = \mathcolor{purple}{\bxtt{0}'}$ 
\Comment{\textcolor{gray}{End of inner diffusion}}
\State $\mathcolor{blue}{\bxtt{t-s^o}} \sim \mathcolor{blue}{q(\bxtt{t-s^o} | \bxzhat, \bxt)}$
\EndFor
\State return $\mathcolor{blue}{\bxz}$
\end{algorithmic}
\end{algorithm} 

\begin{figure}[t]
\centering
    \centerline{\includegraphics[width=\columnwidth]{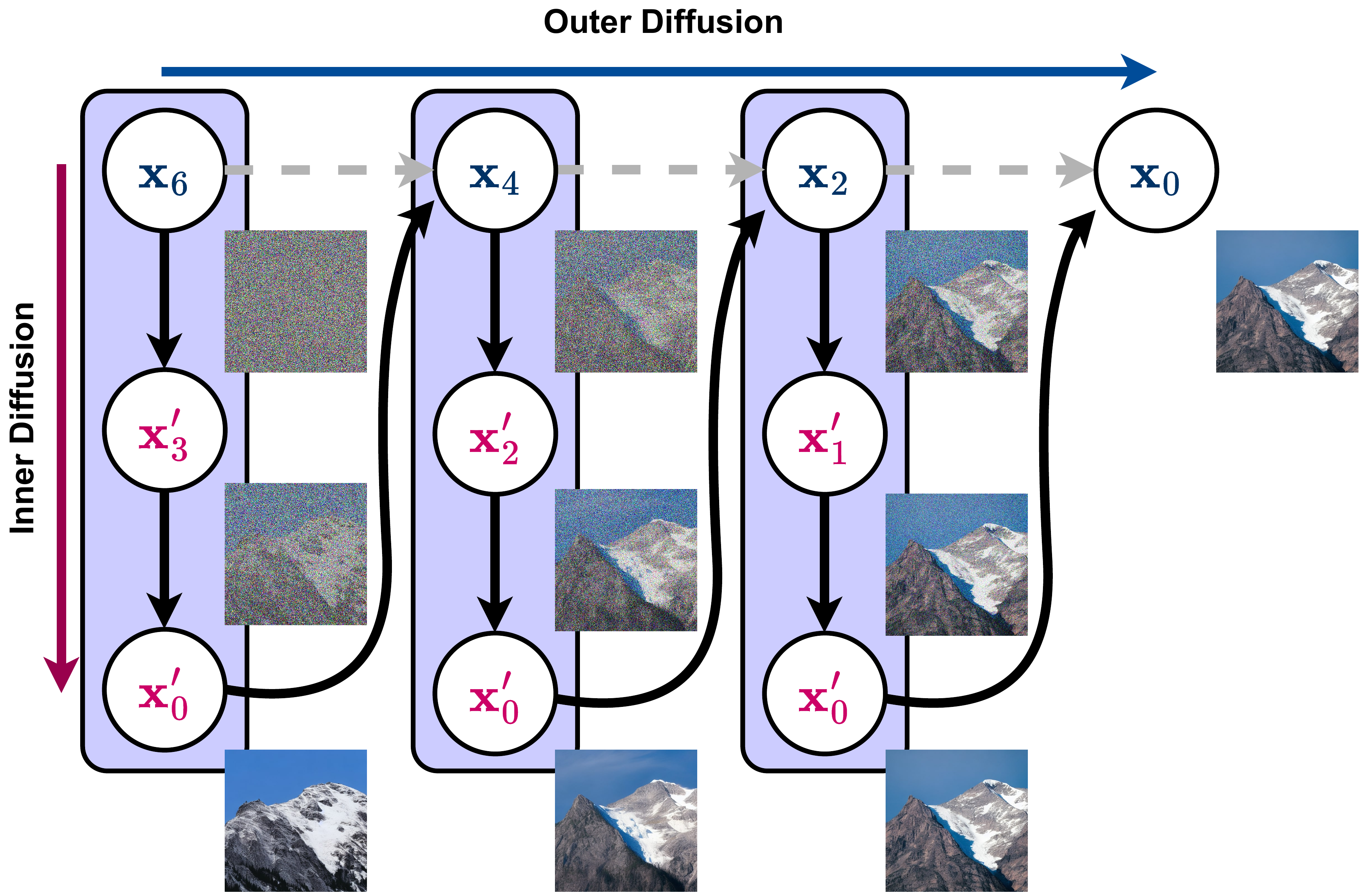}}
    \caption{\textbf{Schematic description of Nested Diffusion.} The outer diffusion process is depicted using the dotted gray arrows}
    \label{fig:diagram}
\end{figure}

We propose a Nested Diffusion process, where an \textbf{outer diffusion} process would utilize the generative sampler $p_\psi(\bxz|\bxt)$ -- itself an \textbf{inner diffusion} process. For simplicity of notation, we denote the outer diffusion variables and distributions in \textcolor{blue}{blue} and the inner diffusion variables and distributions in \textcolor{purple}{purple}. As shown in \Cref{alg:nested-diffusion-sampling} and \Cref{fig:diagram}, for each sampling step $t$ in the outer diffusion, the inner diffusion uses an unaltered (vanilla) DM to generate a plausible image $\mathcolor{blue}{\bxzhat}$, which would then be used to calculate $\mathcolor{blue}{\bxtt{t-s^o}}$ in the outer diffusion. We emphasize that only the inner diffusion uses a DNN. The inner diffusion becomes the outer diffusion's abstraction for a generative model.

Unlike vanilla diffusion processes, Nested Diffusion yields a more detailed $\mathcolor{blue}{\bxzhat}$ at the termination of each outer step. This is because $\mathcolor{blue}{\bxzhat}$ is a sample generated from the multi-step inner diffusion process, and not the mean yielded by a single denoising step. These $\mathcolor{blue}{\bxzhat}$ estimations hint at the final algorithm result while being closer to the image manifold. Using Nested Diffusion, the sampling process becomes an \emph{anytime} algorithm, in which a valid image may be returned if the algorithm is terminated prematurely.

\begin{figure*}
\centering
    \begin{subfigure}[t]{0.33\textwidth}
    \centering\captionsetup{width=0.8\linewidth}
        \includegraphics[width=\columnwidth]{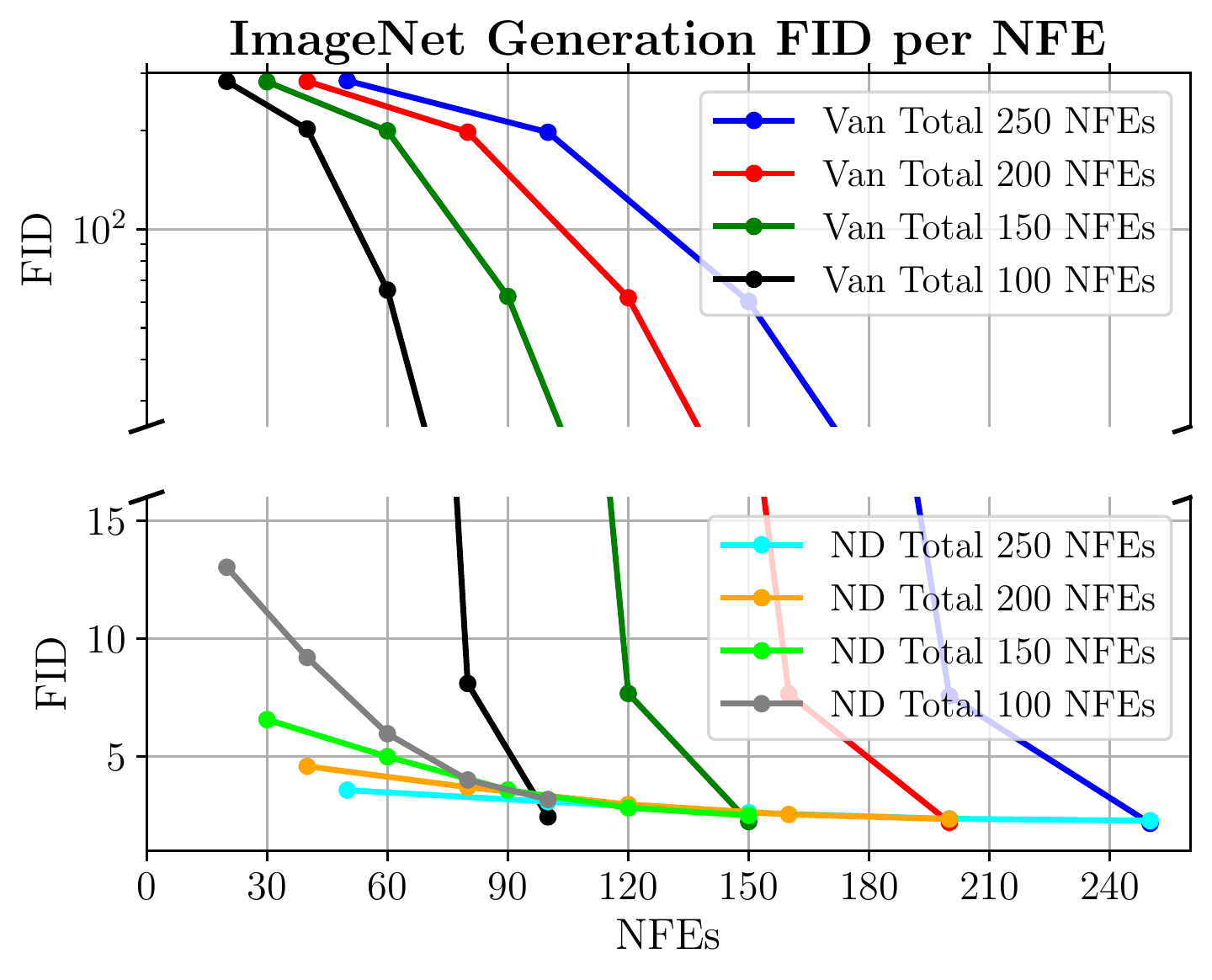}
        \caption{\textbf{50K FID for class-conditional ImageNet generation.} Evaluation of intermediate predictions from Nested (ND) and vanilla (Van) diffusion processes.}
        \label{fig:fid-imagenet-graph}
    \end{subfigure}
    \begin{subfigure}[t]{0.33\textwidth}
        \centering
        \includegraphics[width=\columnwidth]{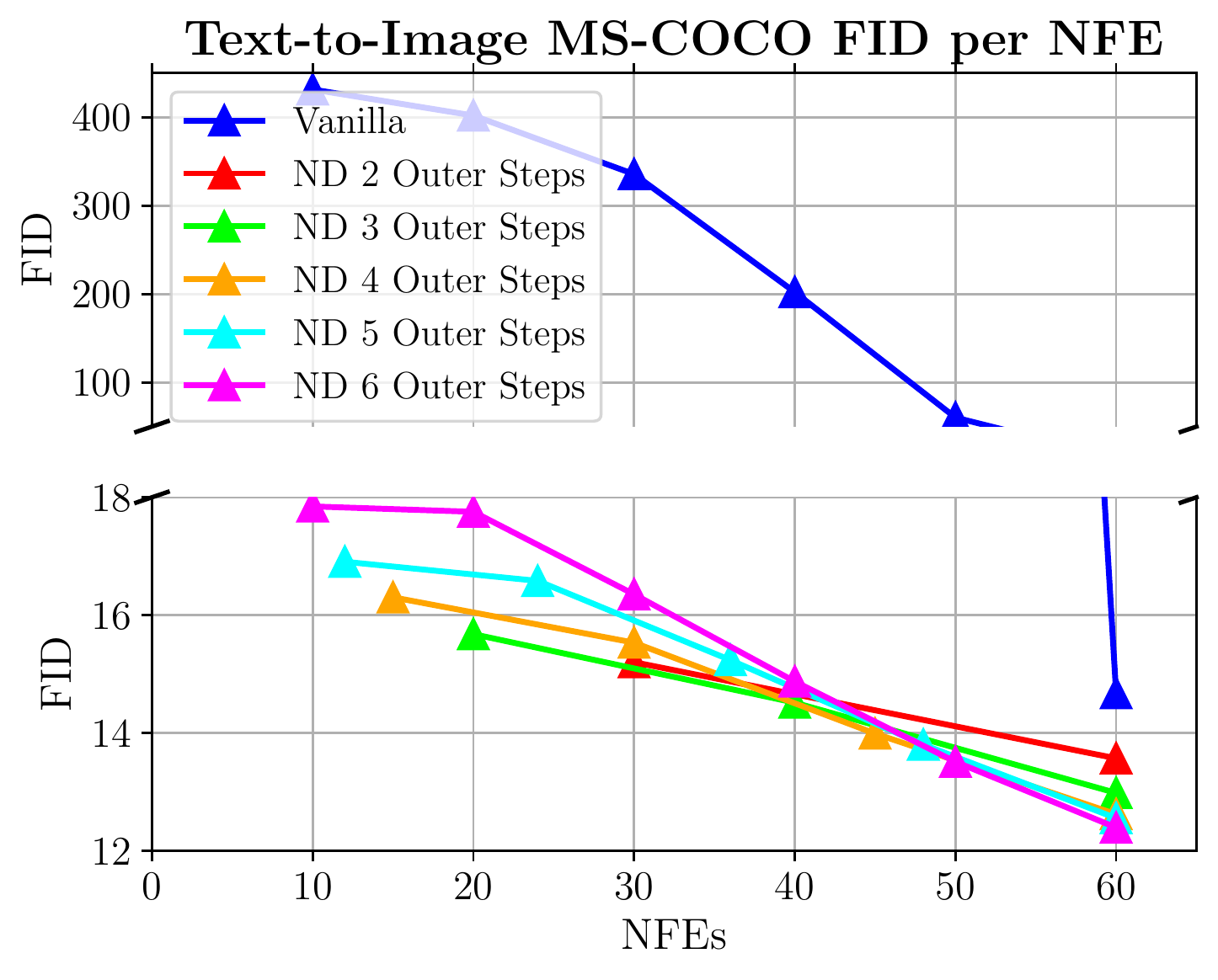}
        \caption{\textbf{MS-COCO 30K FID of text-to-image generation.} Intermediate predictions are generated with a total of 60 NFEs, and different outer step configurations. Vanilla diffusion is equivalent to using Nested Diffusion with one outer step -- in blue.}
        \label{fig:coco-fid}
    \end{subfigure}
    \begin{subfigure}[t]{0.33\textwidth}
        \centering\captionsetup{width=0.8\linewidth}
        \includegraphics[width=\columnwidth]{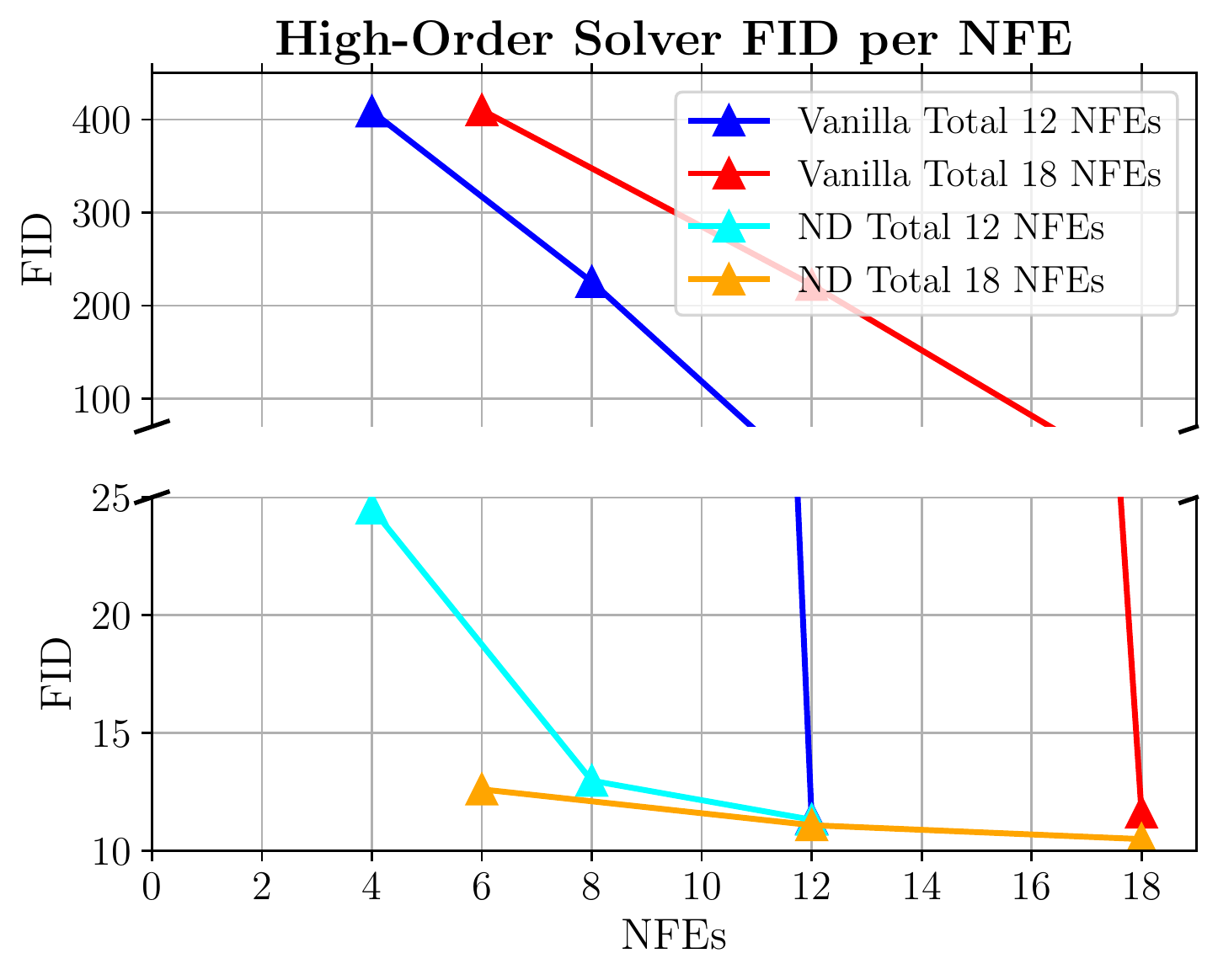}
        \caption{\textbf{MS-COCO 30K FID of text-to-image generation using DPM-Solver++ for the inner diffusion.} Using a high-order solver enables high-quality generation with fewer NFEs.}
        \label{fig:dpm-solver-fid}
    \end{subfigure}
    \vskip -0.05in
    \caption{\textbf{FID as a function of NFE for ImageNet, text-to-image, and high-order solver text-to-image generation.}}
\vskip -0.1in
\end{figure*}

Nested Diffusion requires $|\text{outer steps}| \times |\text{inner steps}|$ NFEs for a complete image genration process. For a given number of NFEs, Nested Diffusion may support any ratio $R_{ND} = \frac{|\text{outer steps}|}{|\text{inner steps}|}$. This ratio represents a tradeoff between fast updates to the predicted image, and the intermediate image quality (see supplementary material).  Additionally, the ratio influences the number of NFEs needed before Nested Diffusion produces its initial intermediate prediction, which occurs at the conclusion of the first inner process. In the extremes, where the number of either outer steps or inner steps is one, the process reverts to vanilla diffusion sampling. In the supplementary material, we propose a metric for comparing between Nested Diffusion with different $R_{ND}$. However, we suggest tuning the $R_{ND}$ parameter according to the specific application and hardware, aiming to provide users with a waiting time between image updates ranging from one to ten seconds -- in Nested Diffusion, the waiting time is the duration of an inner diffusion process.

The computation devoted to each outer step is not required to be the same, i.e.~we can have a different ratio per outer step. As the number of inner steps corresponds to the number of NFEs, changing the length of each outer step determines the computation devoted to this step. In our experiments, we use the same number of inner steps for each outer step for simplicity. We leave for future work to explore dynamic allocation of the number of inner steps per outer step.

\section{Experiments}

    

\begin{figure}[b]
\vskip -0.17in
\centering
    \centerline{\includegraphics[width=0.85\columnwidth]{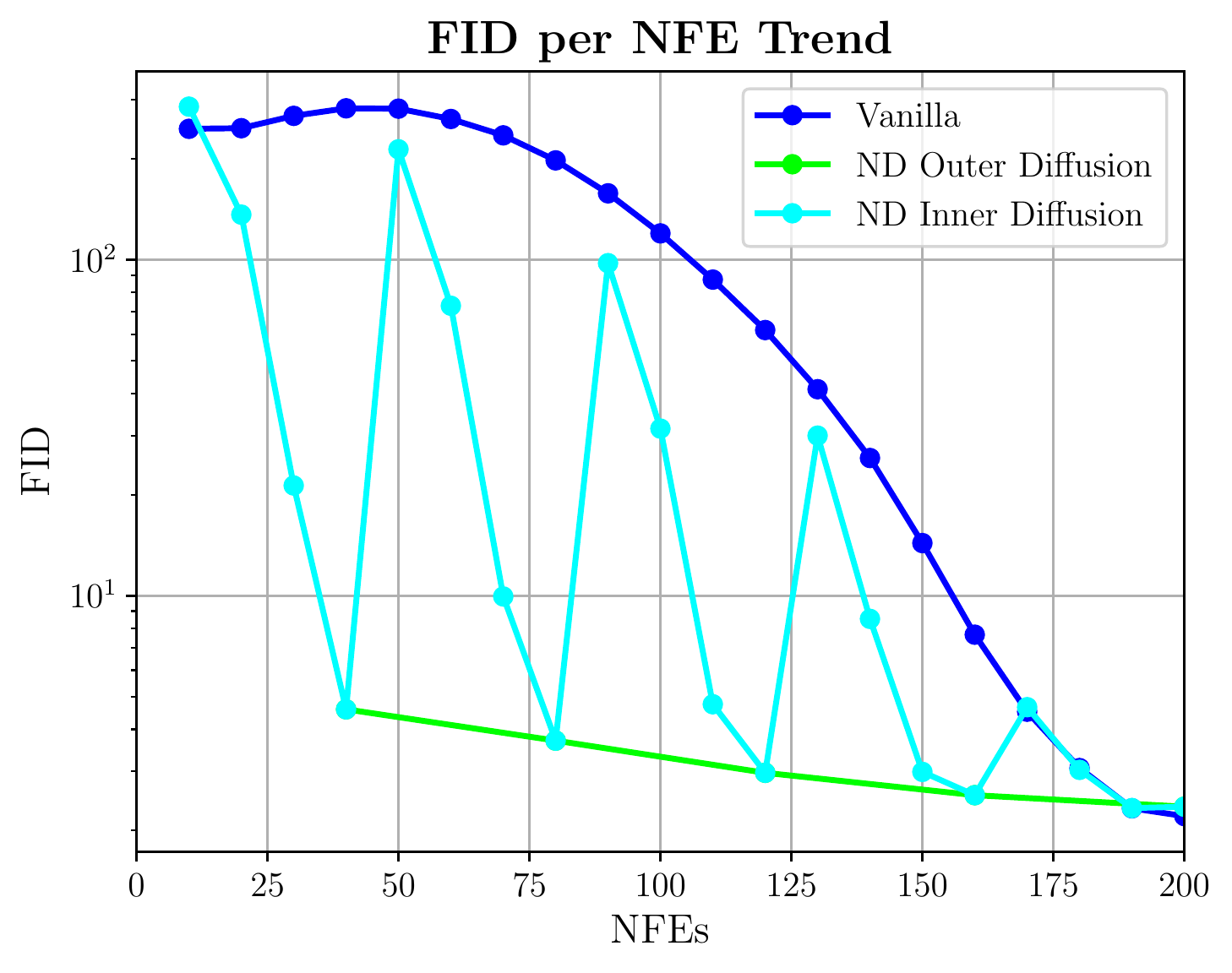}}
    \caption{\textbf{50K FID evaluation of Nested Diffusion's inner and outer diffusion processes.} FID is measured on intermediate predictions of class-conditional ImageNet generation, compared to a vanilla diffusion process,  every 10 NFEs. The Nested Diffusion outer process's FID scores correspond to every fourth inner diffusion measurement, \textit{i.e.}, every 40 NFEs.}
    \label{fig:tradeoff-imagenet}
    
\end{figure}

We evaluate Nested Diffusion as an anytime image generator using a DiT model~\cite{peebles2022scalable} trained on ${256\times256}$-pixel ImageNet~\cite{deng2009imagenet} images and on Stable Diffusion~\cite{latent_diffusion} V1.5. We also show that Nested Diffusion can incorporate an inverse problem solver, and present several examples on CelebA-HQ256~\cite{karras2017progressive} using DDRM~\cite{kawar2022denoising}.
To ensure a fair comparison, we compare Nested Diffusion against the unaltered sampling algorithm (vanilla) using the same DNN models, hyperparameters, and total number of NFEs used. The sampling speed of Nested Diffusion is also equal to that of vanilla diffusion, as sampling time is directly proportional to the total number of NFEs used for generation. 
All experiments use DDIM~\cite{song2020denoising} sampling for the outer diffusion.
The inner diffusion hyperparameters are chosen according to the best practices of the model selected for the experiment. 

    

    

\subsection{Class-Conditional ImageNet Generation}
\label{sec:imagenet}

\begin{figure*}[t]
\centering
    \includegraphics[width=6.4in]{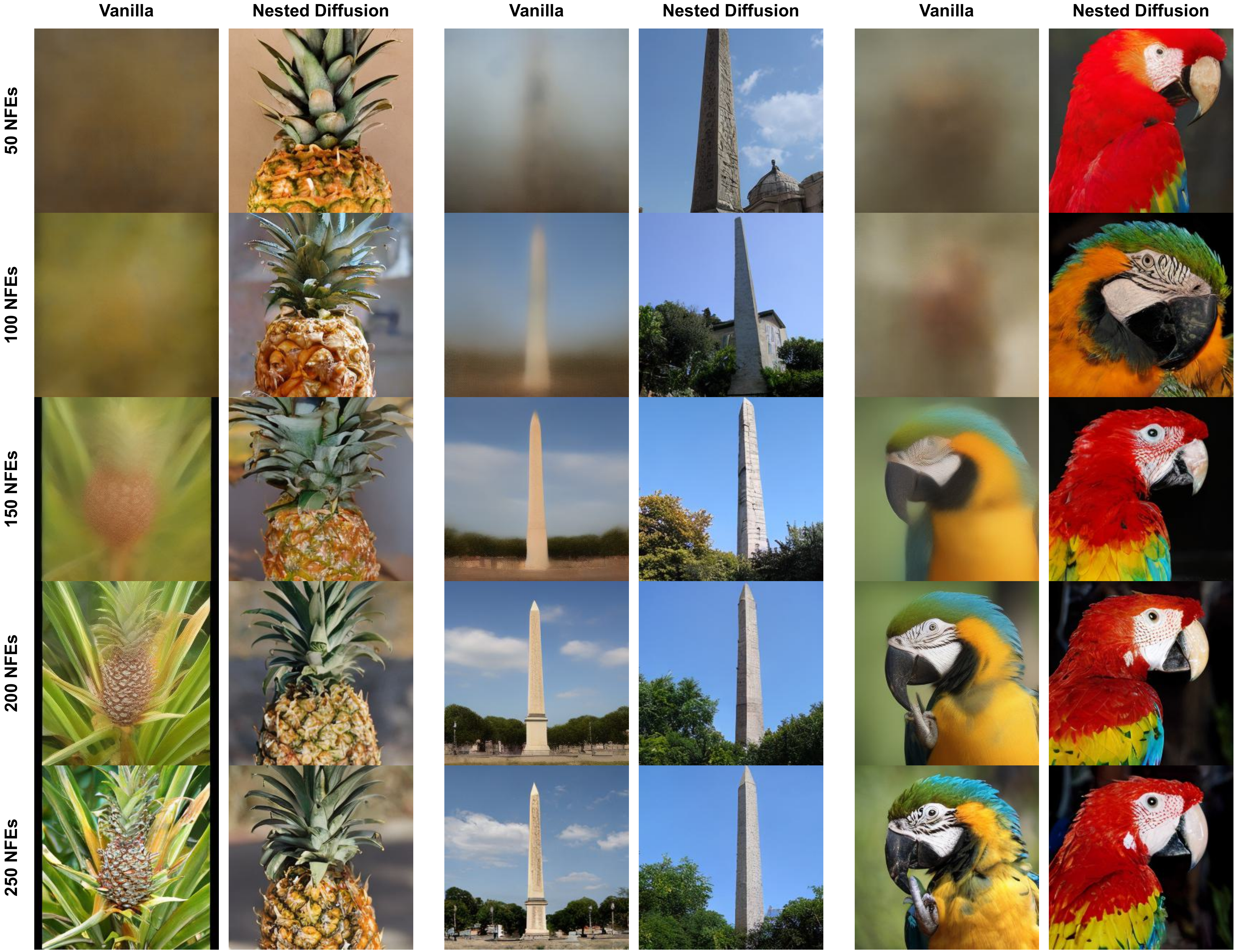}
\caption{\textbf{Samples of class-conditional ImageNet generation, comparing vanilla DMs against Nested Diffusion.}}
\label{fig:imagenet}

\vskip -0.1in
\end{figure*}

The denoising DNN employed in DiT~\cite{peebles2022scalable} uses a VAE~\cite{kingma2013auto} based architecture to decode generated latent samples~\cite{vahdat2021score,latent_diffusion}, thus enabling the application of the DMs in a more efficient latent space. The DNN yields both the mean and variance of a Gaussian distribution $p_\theta(\bxz|\bxt)$, enabling sampling using the reparameterization trick~\cite{kingma2013auto}. In addition, the DNN has been trained with class-labels, using Classifier-Free Guidance (CFG) \cite{ho2022classifier} to generate class-conditional samples. \Cref{fig:imagenet} shows samples generated using $250$ vanilla diffusion steps compared against Nested diffusion with $5$ outer steps and $50$ inner steps each (totaling $250$ NFEs). The latents from the intermediate steps are decoded using the VAE decoder.

In \Cref{fig:fid-imagenet-graph} we compare the FID~\cite{fid} of intermediate estimations of Nested Diffusion with the intermediate estimations of vanilla DMs, for the same number of NFEs\footnote{FID for vanilla diffusion DiT reflect results reproduced by us, which are slightly better than reported in the original paper~\cite{peebles2022scalable}.}. We note that the intermediate FID scores for Nested Diffusion are much better than their vanilla counterparts, while the final result's FID (without early termination) of Nested Diffusion is comparable to the vanilla diffusion.
Exact FID values can be found in the supplementary material.

The sample quality trend for intermediate \textbf{inner samples} $\{\mathcolor{purple}{\hat{\bx}_0'}\}$ is visualized in ~\Cref{fig:tradeoff-imagenet} using FID. The graph shows five distinct drops, corresponding to the five outer diffusion steps. Within each outer step, the inner diffusion's intermediate prediction's quality improves quickly until yielding its final $\mathcolor{purple}{\bxz'}$, which (as shown in \Cref{alg:nested-diffusion-sampling}) is also the outer diffusion's intermediate prediction $\mathcolor{blue}{\bxzhat}$. We observe that the graph bears similarity to simulated annealing with restarts. Nested Diffusion would return the last $\mathcolor{blue}{\bxzhat}$ computed if terminated prematurely -- corresponding to the local minima in the graph, shown in green. 

\begin{figure*}[t]
\centering
    \includegraphics[width=6.0in]{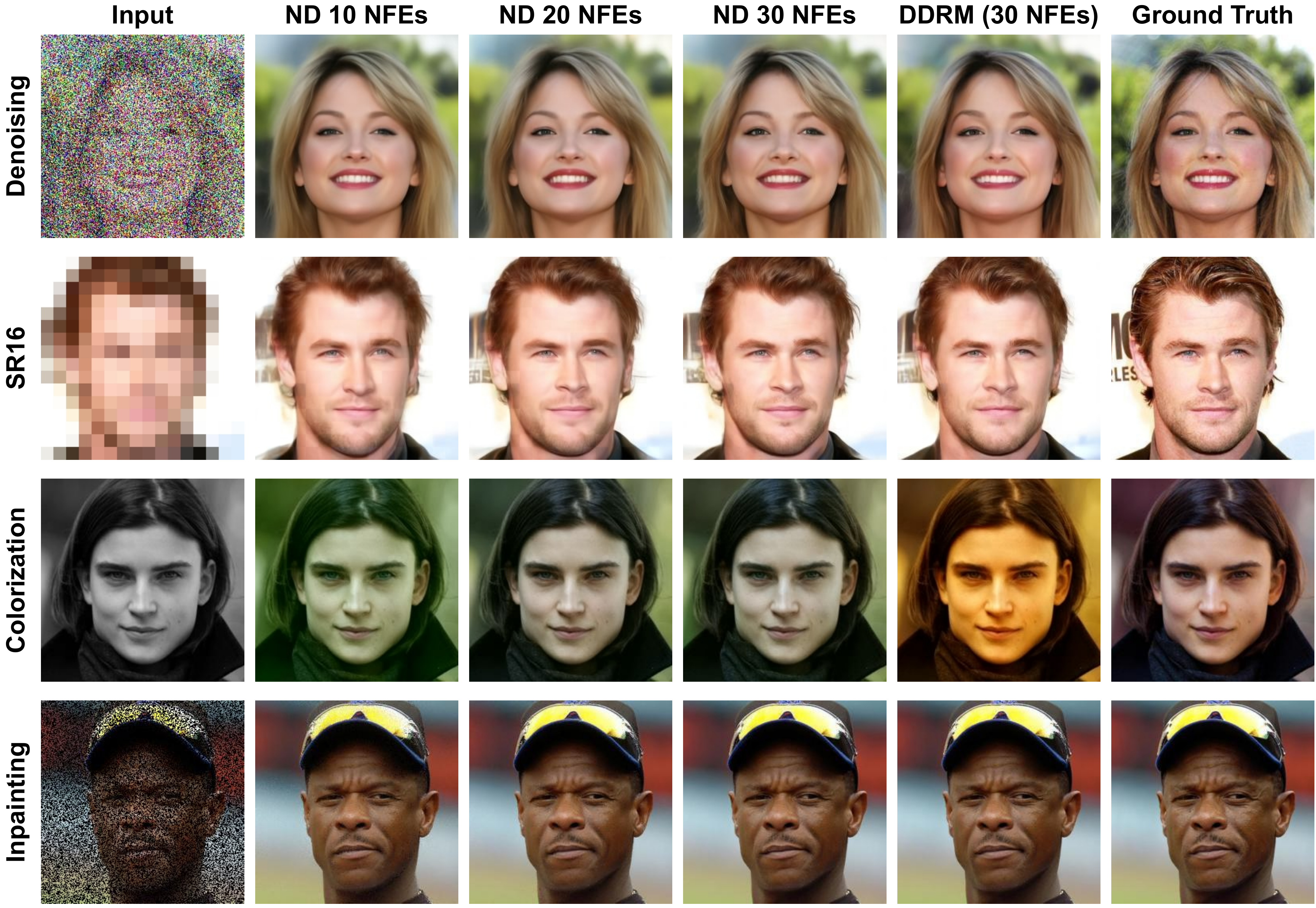}
\caption{\textbf{Inverse problem solutions with Nested Diffusion on CelebA-HQ256 using DDRM.} Nested Diffusion is denoted as ND.}
\label{fig:inverse}

\vskip -0.1in
\end{figure*}

\subsection{Text-to-Image Generation}
\label{sec:stable}
Stable Diffusion is a large text-to-image model capable of generating photo-realistic images for any textual prompt~\cite{latent_diffusion}. We use Stable Diffusion to test Nested Diffusion for text-to-image generation. Similarly to \Cref{sec:imagenet}, Stable Diffusion's process runs in a latent space, and uses CFG~\cite{ho2022classifier} for text-conditional sampling. 
In \Cref{fig:stable}, we present intermediate results from Nested Diffusion using $4$ outer steps and compare them to their counterparts from vanilla Stable Diffusion, decoding intermediate latents using the VAE decoder. The Nested Diffusion sampling process previews satisfactory outputs, highly similar to the end result. The finer details in the images improve with the accumulation of more NFEs. Based on the figure, it is apparent that the intermediate latents obtained from the vanilla DMs do not correspond to valid latents. As a result, when these latents are decoded, they do not produce natural-looking images. 

Following previous work~\cite{latent_diffusion, saharia2022photorealistic, ramesh2022hierarchical, balaji2022ediffi}, we assess Nested Diffusion's performance in text-to-image generation using 30K FID on the MS-COCO~\cite{lin2014microsoft} validation dataset. The results, presented in ~\Cref{fig:coco-fid}, surpass our previous findings, with Nested Diffusion demonstrating comparable intermediate results to vanilla diffusion and slightly improved final results.
More examples for generated images and CLIP-Scores~\cite{hessel2021clipscore} can be found in the supplementary material.

To assess Nested Diffusion's potential and efficiency with advanced high-order schedulers, we replicated the text-to-image experiment while employing DPM-Solver++~\cite{lu2022dpmpp} as the inner diffusion sampling schedule. This change enables using $10$-$20$ NFEs for high quality samples, accelerating generation. As shown in~\Cref{fig:dpm-solver-fid}, Nested Diffusion's final result is of comparable quality to vanilla DM and intermediate prediction quality is improved, demonstrating Nested Diffusion's potential use of high-order solvers.

\subsection{Inverse Problem Solving}

\begin{figure*}[t]
\centering
    \includegraphics[width=6.2in]{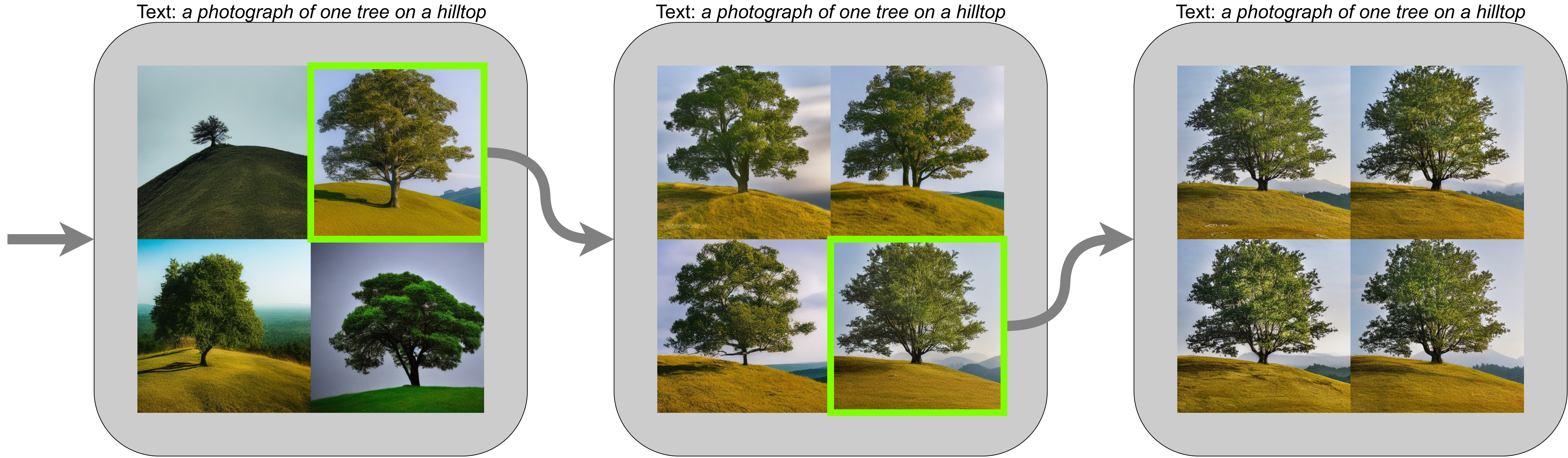}
    \vskip 0.15in
    \includegraphics[width=6.2in]{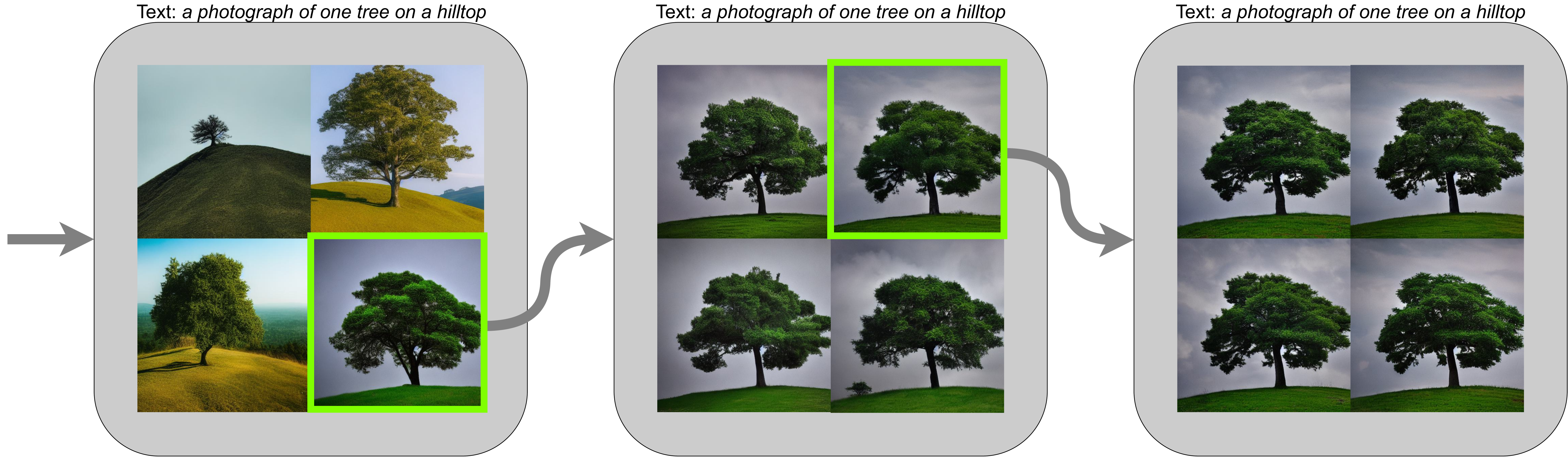}
\caption{\textbf{An example of generation with human feedback.} The top and bottom graphs differ by the user's preference for the image, marked with a bright green frame.}
\label{fig:user-guided}

\vskip -0.1in
\end{figure*}

DMs have demonstrated their effectiveness in tackling inverse problems, whether by training conditional DNNs tailored for specific tasks~\cite{saharia2022palette} or by adapting unconditional DMs DNNs using modified sampling algorithms~\cite{kawar2021stochastic, kawar2022denoising, lugmayr2022repaint, chung2023diffusion, song2023pseudoinverse}.  Following our notation, these inverse problem solvers sample using \Cref{alg:diffusion-sampling}, but exchange the DNN $p_\theta(\bxz|\bxt)$ for a conditional $p_\theta(\bxz|\bxt, \by)$, where $\by$  represents the available measurements. To apply Nested Diffusion in inverse problem solving, a similar substitution is made in the Nested Diffusion sampling \Cref{alg:nested-diffusion-sampling}, where the entire inner diffusion process is replaced with a diffusion-based inverse problem solver conditioned on $\by$. Analogous to image generation scenarios, Nested Diffusion transforms the inverse problem solver into an anytime algorithm, producing plausible results during the sampling process. An exact inverse problem solving algorithm using Nested diffusion is included in the supplementary material.

To evaluate the efficiency of Nested Diffusion for inverse problems, we conduct experiments on the CelebA-HQ256 dataset~\cite{karras2017progressive}, employing DDRM~\cite{kawar2022denoising} as the inverse problem solver. Following DDRM, we rely on a pretrained DDPM~\cite{meng2021sdedit}, and use default hyperparameters except for number of sampling steps used. The results, depicted in Figure \ref{fig:inverse}, demonstrate the generalization capabilities of Nested Diffusion in tackling inverse problems like inpainting, super-resolution, colorization and denoising. The algorithm produces valid intermediate predictions and achieves comparable final results, demonstrating its effectiveness in addressing various inverse problems.

\section{Generation With Human Feedback}
\label{sec:human-feedback}

An emerging area of interest in guided image generation focuses on tuning the generated results to the user's preferences~\cite{zhang2023hive, lee2023aligning}. This type of guidance typically requires user interaction with the model during training, attempting to fine-tune the DMs's generation process using direct feedback. The fine-tuned models show a greater capability to match the model's behaviour with the user's demands. 

Nested Diffusion, by its inherent design, allows users to view the generated output throughout the sampling algorithm, enabling straightforward guidance of the process towards desired outcomes. For instance, in many cases multiple images are generated simultaneously using different random noise vectors, to provide the user with several alternative results. If Nested Diffusion is used for sampling, a user can see a likeness of the final possibilities. By pruning unwanted generation attempts, computational resources can be efficiently allocated to explore additional options based on the remaining intermediate predictions. 

In contrast with model fine-tuning methods, Nested Diffusion can incorporate human feedback inherently, with no requirement for further training.  Moreover, Nested Diffusion may be combined with fine-tuned DMs to further enhance their consistency with the user's preferences.

~\Cref{fig:user-guided} shows an example of a human feedback-based generation scheme implemented using Nested Diffusion. The samples were generated following the generation details provided in ~\Cref{sec:stable}, using $3$ outer steps with $20$ inner steps each.  At the conclusion of each inner diffusion process, the user is presented with four intermediate samples, allowing them to select their desired output. The chosen sample is then propagated to replace the other samples, and the sampling algorithm resumes its execution.

\begin{figure*}[t]
\centering
    \includegraphics[width=6.2in]{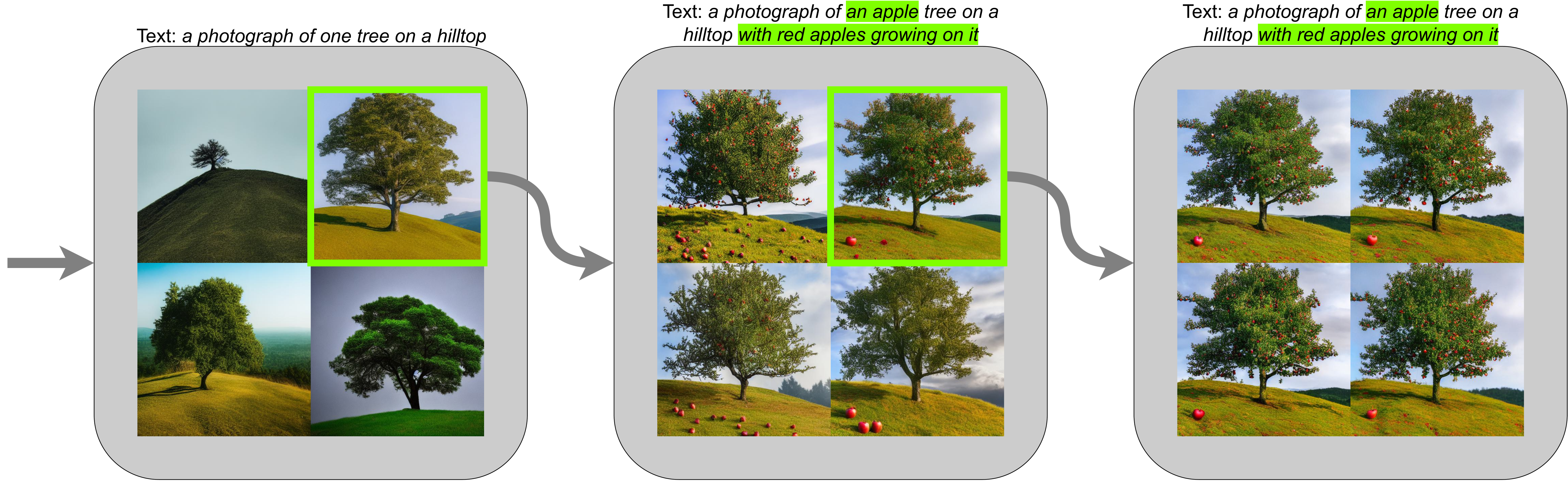}
    \vskip 0.15in
    \includegraphics[width=6.2in]{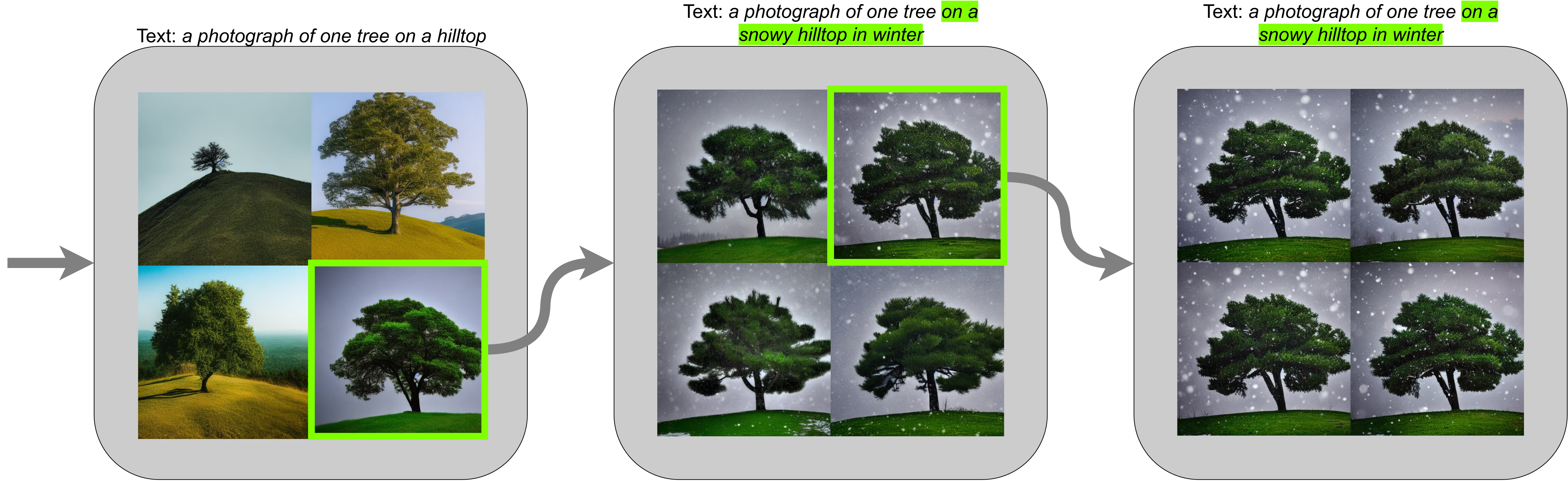}
\caption{\textbf{An example of interactive content creation using human feedback.} The image selected by the user in each graph is marked with a bright green frame. The text prompt is changed after the first outer step. }
\label{fig:user-editing}

\vskip -0.1in
\end{figure*}

In addition to selecting from a pool of several options to guide the generation, further refinement of the sampling procedure can be achieved by integrating editing techniques into the sampling process. This editing can be accomplished using one of the many available diffusion-based image editing methods ~\cite{meng2021sdedit, avrahami2022blended, hertz2022prompt, brooks2022instructpix2pix, kawar2023imagic} in tandem with Nested Diffusion, by modifying  the intermediate $\mathcolor{blue}{\bxzhat}$, similar to SDEdit~\cite{meng2021sdedit}, or adjusting the subsequent inner diffusion process. However, we adopt a simpler approach: we add details to the textual prompt at the conclusion of each inner diffusion process during text-to-image generation. In ~\Cref{fig:user-editing}, we show some promising results for our approach.

\section{Related Work}
The noise scheduling in reverse diffusion sampling has garnered considerable attention in recent years~\cite{chen2023importance}. DDPM~\cite{ho2020denoising} implements a linearly increasing schedule, while IDPM~\cite{nichol2021improved} demonstrates the potential of cosine scheduling in achieving improved sampling outcomes. In DDIM~\cite{song2020denoising}, the authors eliminate the forward diffusion's Markovian assumption, resulting in a deterministic reverse process that can accelerate sampling. Using ODE solving methods~\cite{liu2022pseudo, lu2022dpm, lu2022dpmpp}, the sampling process can attain superior results and faster generation.
Nested Diffusion, while not strictly a noise schedule, intertwines two separate noise schedules (the inner and outer diffusion processes) into one sampling process.

Creative scheduling of noise can be employed in other domains besides image generation. In the field of image editing, SDEdit~\cite{meng2021sdedit} degrades an edited clean image with noise and subsequently denoises it using a DM. This process enhances the realism of the edited image, facilitating photo-realistic editing using simple tools. 


Noise has also been used in inverse problem solvers to ``time-travel'' in the diffusion process~\cite{lugmayr2022repaint, wang2022zero}. These approaches revert the diffusion process to a previous step by adding random Gaussian noise, requiring additional NFEs and enhancing image fidelity. However, unlike Nested Diffusion, these methods add noise to revert a specific number of steps (a hyperparameter) and do not involve multiple diffusion processes. Consequently, they do not benefit from the anytime algorithm property and require more NFEs compared to alternative approaches.

Nested Diffusion is orthogonal to many diffusion acceleration methods, such as the fast sampling offered by DPM-Solver++~\cite{lu2022dpmpp} shown in~\Cref{sec:stable}, and may work well with parallelized sampling~\cite{pokle2022deep, shih2023parallel} or trajectory-based methods~\cite{liu2022flow, song2023consistency}. In this work, we have not delved into some of these avenues for several reasons; Parallelized sampling typically requires more NFEs per image even when reducing overall sampling speed. Trajectory-based methods, while requiring a fraction of the resources, require additional training and may still fall short of achieving the performance standards set by multi-step diffusion sampling techniques~\cite{latent_diffusion, peebles2022scalable}. We have chosen to optimize Nested Diffusion for high quality at the expense of multiple steps and use NFEs to measure our computational resources. Nevertheless, combining these methods with anytime generation holds promise for future work.

\vfill
\section{Conclusion}
We introduced Nested Diffusion, a probabilistic approach that harnesses a diffusion process as a building block in another diffusion process. Our approach allows anytime sampling from a pre-trained diffusion model. Through quantitative and qualitative evaluation, we demonstrated the effectiveness of Nested Diffusion in tandem with state-of-the-art DMs, including latent diffusion, CFG-based class-conditional generation, and text-to-image generation. Furthermore, we explored the potential of Nested Diffusion in enabling generation with human feedback and facilitating interactive content creation. Our findings highlight the versatility and practical applications of Nested Diffusion in various domains of generative modeling.
\section{Acknowledgements}
This work was supported by the Israel Science Foundation grant 2318/22, the Ollendorff Minerva Center, Technion, a gift from KLA, and the Council For Higher Education - Planning \& Budgeting Committee, Israel. 

\nocite{hessel2021clipscore, lpips, fu2023learning}

\newpage 

{\small
\bibliographystyle{ieee_fullname}
\bibliography{refs}

\begin{thebibliography}{10}\itemsep=-1pt

\bibitem{avrahami2022blended}
Omri Avrahami, Dani Lischinski, and Ohad Fried.
\newblock Blended diffusion for text-driven editing of natural images.
\newblock In {\em Proceedings of the IEEE/CVF Conference on Computer Vision and
  Pattern Recognition (CVPR)}, pages 18208--18218, June 2022.

\bibitem{balaji2022ediffi}
Yogesh Balaji, Seungjun Nah, Xun Huang, Arash Vahdat, Jiaming Song, Karsten
  Kreis, Miika Aittala, Timo Aila, Samuli Laine, Bryan Catanzaro, et~al.
\newblock ediffi: Text-to-image diffusion models with an ensemble of expert
  denoisers.
\newblock {\em arXiv preprint arXiv:2211.01324}, 2022.

\bibitem{boddy1989solving}
Mark Boddy and Thomas~L Dean.
\newblock {\em Solving time-dependent planning problems}.
\newblock Brown University, Department of Computer Science, 1989.

\bibitem{brooks2022instructpix2pix}
Tim Brooks, Aleksander Holynski, and Alexei~A. Efros.
\newblock Instructpix2pix: Learning to follow image editing instructions.
\newblock In {\em CVPR}, 2023.

\bibitem{chen2023importance}
Ting Chen.
\newblock On the importance of noise scheduling for diffusion models.
\newblock {\em arXiv preprint arXiv:2301.10972}, 2023.

\bibitem{chung2023diffusion}
Hyungjin Chung, Jeongsol Kim, Michael~Thompson Mccann, Marc~Louis Klasky, and
  Jong~Chul Ye.
\newblock Diffusion posterior sampling for general noisy inverse problems.
\newblock In {\em The Eleventh International Conference on Learning
  Representations}, 2023.

\bibitem{chung2022score}
Hyungjin Chung and Jong~Chul Ye.
\newblock Score-based diffusion models for accelerated {MRI}.
\newblock {\em Medical Image Analysis}, 80:102479, 2022.

\bibitem{deng2009imagenet}
Jia Deng, Wei Dong, Richard Socher, Li-Jia Li, Kai Li, and Li Fei-Fei.
\newblock Imagenet: A large-scale hierarchical image database.
\newblock In {\em 2009 IEEE conference on computer vision and pattern
  recognition}, pages 248--255. Ieee, 2009.

\bibitem{imagenet}
Jia Deng, Wei Dong, Richard Socher, Li-Jia Li, Kai Li, and Li Fei-Fei.
\newblock {ImageNet: A large-scale hierarchical image database}.
\newblock In {\em 2009 IEEE Conference on Computer Vision and Pattern
  Recognition}, pages 248--255, 2009.

\bibitem{dhariwal2021diffusion}
Prafulla Dhariwal and Alexander Nichol.
\newblock Diffusion models beat {GANs} on image synthesis.
\newblock {\em Advances in Neural Information Processing Systems},
  34:8780--8794, 2021.

\bibitem{fu2023learning}
Stephanie Fu*, Netanel Tamir*, Shobhita Sundaram*, Lucy Chai, Richard Zhang,
  Tali Dekel, and Phillip Isola.
\newblock Dreamsim: Learning new dimensions of human visual similarity using
  synthetic data.
\newblock {\em arXiv:2306.09344}, 2023.

\bibitem{goodfellow2014generative}
Ian Goodfellow, Jean Pouget-Abadie, Mehdi Mirza, Bing Xu, David Warde-Farley,
  Sherjil Ozair, Aaron Courville, and Yoshua Bengio.
\newblock Generative adversarial nets.
\newblock In {\em Advances in neural information processing systems}, pages
  2672--2680, 2014.

\bibitem{grass1996anytime}
Joshua Grass and Shlomo Zilberstein.
\newblock Anytime algorithm development tools.
\newblock {\em ACM SIGART Bulletin}, 7(2):20--27, 1996.

\bibitem{hertz2022prompt}
Amir Hertz, Ron Mokady, Jay Tenenbaum, Kfir Aberman, Yael Pritch, and Daniel
  Cohen-Or.
\newblock Prompt-to-prompt image editing with cross attention control.
\newblock {\em arXiv preprint arXiv:2208.01626}, 2022.

\bibitem{hessel2021clipscore}
Jack Hessel, Ari Holtzman, Maxwell Forbes, Ronan~Le Bras, and Yejin Choi.
\newblock Clipscore: A reference-free evaluation metric for image captioning.
\newblock {\em arXiv preprint arXiv:2104.08718}, 2021.

\bibitem{fid}
Martin Heusel, Hubert Ramsauer, Thomas Unterthiner, Bernhard Nessler, and Sepp
  Hochreiter.
\newblock Gans trained by a two time-scale update rule converge to a local nash
  equilibrium.
\newblock In {\em Advances in Neural Information Processing Systems},
  volume~30, 2017.

\bibitem{ho2020denoising}
Jonathan Ho, Ajay Jain, and Pieter Abbeel.
\newblock Denoising diffusion probabilistic models.
\newblock {\em Advances in Neural Information Processing Systems},
  33:6840--6851, 2020.

\bibitem{ho2022classifier}
Jonathan Ho and Tim Salimans.
\newblock Classifier-free diffusion guidance.
\newblock {\em arXiv preprint arXiv:2207.12598}, 2022.

\bibitem{horsch2013anytime}
Michael~C Horsch and David~L Poole.
\newblock An anytime algorithm for decision making under uncertainty.
\newblock {\em arXiv preprint arXiv:1301.7384}, 2013.

\bibitem{jeong2021diff}
Myeonghun Jeong, Hyeongju Kim, Sung~Jun Cheon, Byoung~Jin Choi, and Nam~Soo
  Kim.
\newblock Diff-tts: A denoising diffusion model for text-to-speech.
\newblock {\em arXiv preprint arXiv:2104.01409}, 2021.

\bibitem{karras2017progressive}
Tero Karras, Timo Aila, Samuli Laine, and Jaakko Lehtinen.
\newblock Progressive growing of gans for improved quality, stability, and
  variation.
\newblock {\em arXiv preprint arXiv:1710.10196}, 2017.

\bibitem{kawar2022denoising}
Bahjat Kawar, Michael Elad, Stefano Ermon, and Jiaming Song.
\newblock Denoising diffusion restoration models.
\newblock In {\em Advances in Neural Information Processing Systems}, 2022.

\bibitem{kawar2023gsure}
Bahjat Kawar, Noam Elata, Tomer Michaeli, and Michael Elad.
\newblock Gsure-based diffusion model training with corrupted data.
\newblock {\em arXiv preprint arXiv:2305.13128}, 2023.

\bibitem{kawar2021stochastic}
Bahjat Kawar, Gregory Vaksman, and Michael Elad.
\newblock Stochastic image denoising by sampling from the posterior
  distribution.
\newblock In {\em Proceedings of the IEEE/CVF International Conference on
  Computer Vision}, pages 1866--1875, 2021.

\bibitem{kawar2023imagic}
Bahjat Kawar, Shiran Zada, Oran Lang, Omer Tov, Huiwen Chang, Tali Dekel, Inbar
  Mosseri, and Michal Irani.
\newblock Imagic: Text-based real image editing with diffusion models.
\newblock In {\em Conference on Computer Vision and Pattern Recognition 2023},
  2023.

\bibitem{kingma2013auto}
Diederik~P Kingma and Max Welling.
\newblock Auto-encoding variational bayes.
\newblock {\em arXiv preprint arXiv:1312.6114}, 2013.

\bibitem{kong2020diffwave}
Zhifeng Kong, Wei Ping, Jiaji Huang, Kexin Zhao, and Bryan Catanzaro.
\newblock Diffwave: A versatile diffusion model for audio synthesis.
\newblock {\em arXiv preprint arXiv:2009.09761}, 2020.

\bibitem{lee2023aligning}
Kimin Lee, Hao Liu, Moonkyung Ryu, Olivia Watkins, Yuqing Du, Craig Boutilier,
  Pieter Abbeel, Mohammad Ghavamzadeh, and Shixiang~Shane Gu.
\newblock Aligning text-to-image models using human feedback.
\newblock {\em arXiv preprint arXiv:2302.12192}, 2023.

\bibitem{lin2014microsoft}
Tsung-Yi Lin, Michael Maire, Serge Belongie, James Hays, Pietro Perona, Deva
  Ramanan, Piotr Doll{\'a}r, and C~Lawrence Zitnick.
\newblock Microsoft coco: Common objects in context.
\newblock In {\em Computer Vision--ECCV 2014: 13th European Conference, Zurich,
  Switzerland, September 6-12, 2014, Proceedings, Part V 13}, pages 740--755.
  Springer, 2014.

\bibitem{liu2022pseudo}
Luping Liu, Yi Ren, Zhijie Lin, and Zhou Zhao.
\newblock Pseudo numerical methods for diffusion models on manifolds.
\newblock {\em arXiv preprint arXiv:2202.09778}, 2022.

\bibitem{liu2022flow}
Xingchao Liu, Chengyue Gong, and Qiang Liu.
\newblock Flow straight and fast: Learning to generate and transfer data with
  rectified flow.
\newblock {\em arXiv preprint arXiv:2209.03003}, 2022.

\bibitem{lu2022dpm}
Cheng Lu, Yuhao Zhou, Fan Bao, Jianfei Chen, Chongxuan Li, and Jun Zhu.
\newblock Dpm-solver: A fast ode solver for diffusion probabilistic model
  sampling in around 10 steps.
\newblock {\em arXiv preprint arXiv:2206.00927}, 2022.

\bibitem{lu2022dpmpp}
Cheng Lu, Yuhao Zhou, Fan Bao, Jianfei Chen, Chongxuan Li, and Jun Zhu.
\newblock Dpm-solver++: Fast solver for guided sampling of diffusion
  probabilistic models.
\newblock {\em arXiv preprint arXiv:2211.01095}, 2022.

\bibitem{lugmayr2022repaint}
Andreas Lugmayr, Martin Danelljan, Andres Romero, Fisher Yu, Radu Timofte, and
  Luc Van~Gool.
\newblock Repaint: Inpainting using denoising diffusion probabilistic models.
\newblock In {\em Proceedings of the IEEE/CVF Conference on Computer Vision and
  Pattern Recognition}, pages 11461--11471, 2022.

\bibitem{meng2021sdedit}
Chenlin Meng, Yutong He, Yang Song, Jiaming Song, Jiajun Wu, Jun-Yan Zhu, and
  Stefano Ermon.
\newblock Sdedit: Guided image synthesis and editing with stochastic
  differential equations.
\newblock In {\em International Conference on Learning Representations}, 2021.

\bibitem{nichol2021improved}
Alexander~Quinn Nichol and Prafulla Dhariwal.
\newblock Improved denoising diffusion probabilistic models.
\newblock In {\em International Conference on Machine Learning}, pages
  8162--8171. PMLR, 2021.

\bibitem{peebles2022scalable}
William Peebles and Saining Xie.
\newblock Scalable diffusion models with transformers.
\newblock {\em arXiv preprint arXiv:2212.09748}, 2022.

\bibitem{pokle2022deep}
Ashwini Pokle, Zhengyang Geng, and J~Zico Kolter.
\newblock Deep equilibrium approaches to diffusion models.
\newblock {\em Advances in Neural Information Processing Systems},
  35:37975--37990, 2022.

\bibitem{qiao2022dynamic}
Zhuoran Qiao, Weili Nie, Arash Vahdat, Thomas~F Miller~III, and Anima
  Anandkumar.
\newblock Dynamic-backbone protein-ligand structure prediction with multiscale
  generative diffusion models.
\newblock {\em arXiv preprint arXiv:2209.15171}, 2022.

\bibitem{ramesh2022hierarchical}
Aditya Ramesh, Prafulla Dhariwal, Alex Nichol, Casey Chu, and Mark Chen.
\newblock Hierarchical text-conditional image generation with clip latents.
\newblock {\em arXiv preprint arXiv:2204.06125}, 2022.

\bibitem{latent_diffusion}
Robin Rombach, Andreas Blattmann, Dominik Lorenz, Patrick Esser, and Bj\"orn
  Ommer.
\newblock High-resolution image synthesis with latent diffusion models.
\newblock In {\em Proceedings of the IEEE/CVF Conference on Computer Vision and
  Pattern Recognition (CVPR)}, pages 10684--10695, June 2022.

\bibitem{saharia2022palette}
Chitwan Saharia, William Chan, Huiwen Chang, Chris Lee, Jonathan Ho, Tim
  Salimans, David Fleet, and Mohammad Norouzi.
\newblock Palette: Image-to-image diffusion models.
\newblock In {\em ACM SIGGRAPH 2022 Conference Proceedings}, pages 1--10, 2022.

\bibitem{saharia2022photorealistic}
Chitwan Saharia, William Chan, Saurabh Saxena, Lala Li, Jay Whang, Emily~L
  Denton, Kamyar Ghasemipour, Raphael Gontijo~Lopes, Burcu Karagol~Ayan, Tim
  Salimans, et~al.
\newblock Photorealistic text-to-image diffusion models with deep language
  understanding.
\newblock {\em Advances in Neural Information Processing Systems},
  35:36479--36494, 2022.

\bibitem{salimans2022progressive}
Tim Salimans and Jonathan Ho.
\newblock Progressive distillation for fast sampling of diffusion models.
\newblock In {\em International Conference on Learning Representations}, 2022.

\bibitem{schneuing2022structure}
Arne Schneuing, Yuanqi Du, Charles Harris, Arian Jamasb, Ilia Igashov, Weitao
  Du, Tom Blundell, Pietro Li{\'o}, Carla Gomes, Max Welling, Michael
  Bronstein, and Bruno Correia.
\newblock Structure-based drug design with equivariant diffusion models.
\newblock {\em arXiv preprint arXiv:2210.13695}, 2022.

\bibitem{shih2023parallel}
Andy Shih, Suneel Belkhale, Stefano Ermon, Dorsa Sadigh, and Nima Anari.
\newblock Parallel sampling of diffusion models.
\newblock {\em arXiv preprint arXiv:2305.16317}, 2023.

\bibitem{sohl2015deep}
Jascha Sohl-Dickstein, Eric Weiss, Niru Maheswaranathan, and Surya Ganguli.
\newblock Deep unsupervised learning using nonequilibrium thermodynamics.
\newblock In {\em International Conference on Machine Learning}, pages
  2256--2265. PMLR, 2015.

\bibitem{song2020denoising}
Jiaming Song, Chenlin Meng, and Stefano Ermon.
\newblock Denoising diffusion implicit models.
\newblock In {\em International Conference on Learning Representations}, 2020.

\bibitem{song2023pseudoinverse}
Jiaming Song, Arash Vahdat, Morteza Mardani, and Jan Kautz.
\newblock Pseudoinverse-guided diffusion models for inverse problems.
\newblock In {\em International Conference on Learning Representations (ICLR)},
  May 2023.

\bibitem{song2023consistency}
Yang Song, Prafulla Dhariwal, Mark Chen, and Ilya Sutskever.
\newblock Consistency models.
\newblock {\em arXiv preprint arXiv:2303.01469}, 2023.

\bibitem{song2019generative}
Yang Song and Stefano Ermon.
\newblock Generative modeling by estimating gradients of the data distribution.
\newblock {\em Advances in Neural Information Processing Systems}, 32, 2019.

\bibitem{song2023solving}
Yang Song, Liyue Shen, Lei Xing, and Stefano Ermon.
\newblock Solving inverse problems in medical imaging with score-based
  generative models.
\newblock In {\em International Conference on Learning Representations}, 2023.

\bibitem{vahdat2021score}
Arash Vahdat, Karsten Kreis, and Jan Kautz.
\newblock Score-based generative modeling in latent space.
\newblock {\em Advances in Neural Information Processing Systems},
  34:11287--11302, 2021.

\bibitem{wang2022zero}
Yinhuai Wang, Jiwen Yu, and Jian Zhang.
\newblock Zero-shot image restoration using denoising diffusion null-space
  model.
\newblock {\em arXiv preprint arXiv:2212.00490}, 2022.

\bibitem{xiao2021tackling}
Zhisheng Xiao, Karsten Kreis, and Arash Vahdat.
\newblock Tackling the generative learning trilemma with denoising diffusion
  {GAN}s.
\newblock In {\em International Conference on Learning Representations (ICLR)},
  2022.

\bibitem{lpips}
Richard Zhang, Phillip Isola, Alexei~A Efros, Eli Shechtman, and Oliver Wang.
\newblock The unreasonable effectiveness of deep features as a perceptual
  metric.
\newblock In {\em CVPR}, 2018.

\bibitem{zhang2023hive}
Shu Zhang, Xinyi Yang, Yihao Feng, Can Qin, Chia-Chih Chen, Ning Yu, Zeyuan
  Chen, Huan Wang, Silvio Savarese, Stefano Ermon, et~al.
\newblock Hive: Harnessing human feedback for instructional visual editing.
\newblock {\em arXiv preprint arXiv:2303.09618}, 2023.

\bibitem{zilberstein1996using}
Shlomo Zilberstein.
\newblock Using anytime algorithms in intelligent systems.
\newblock {\em AI magazine}, 17(3):73--73, 1996.

\end{thebibliography}
}

\appendix

\section{Outer Steps -- Inner Steps Trade-off}
\label{sec:tradeoff}

\begin{figure*}
\centering
    \centerline{\includegraphics[width=5.9in]{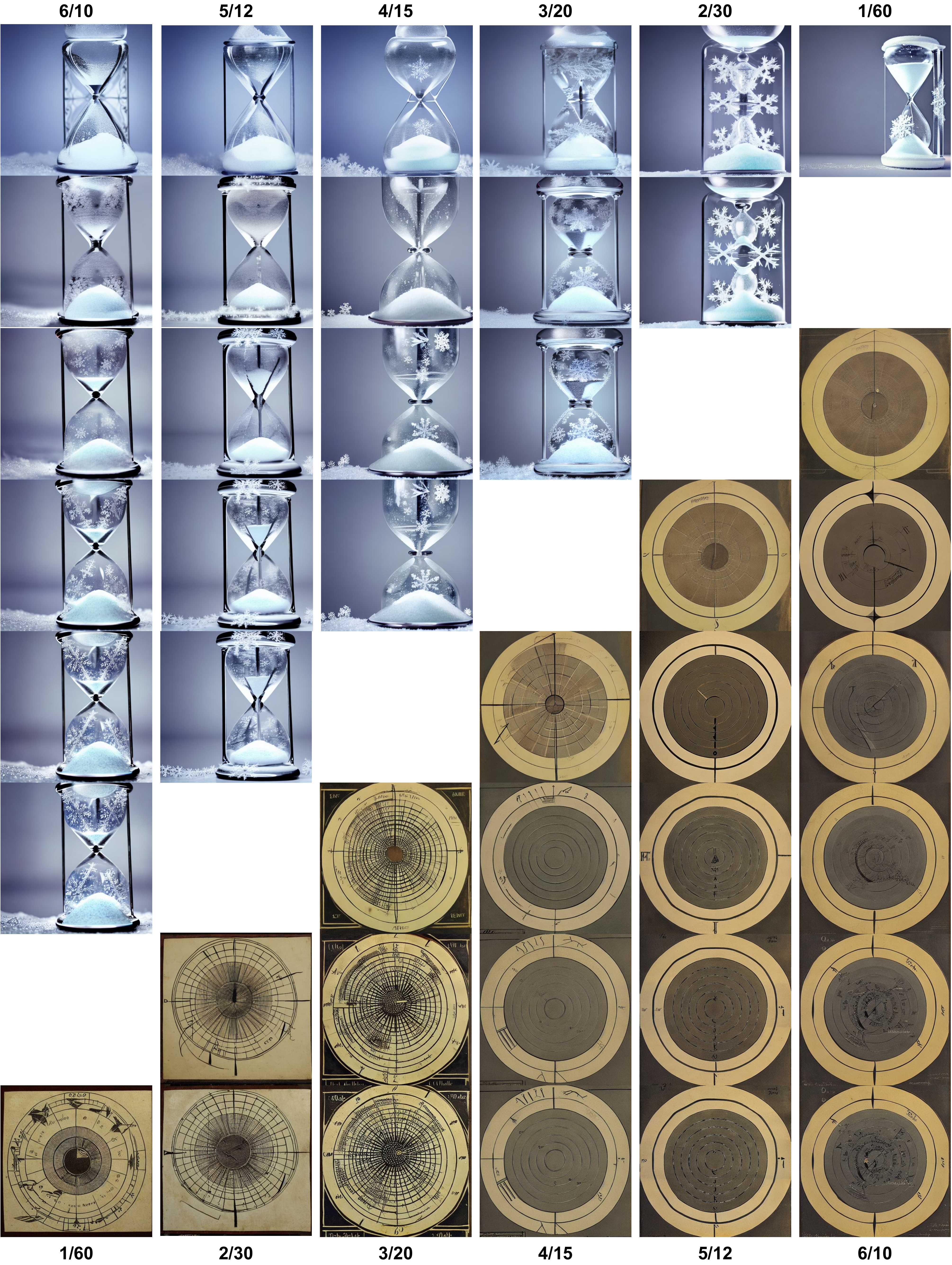}}
    \caption{\textbf{Qualitative examples of Nested Diffusion with different ratios $R_{ND}$.} Each column denoted with $|\text{outer steps}|$/$|\text{inner steps}|$ at the top or bottom. Top text: \textit{a photograph of an hourglass filled with snowflakes}. Bottom text: \textit{a diagram of an ancient sundial}. Diffusion process progresses from top to bottom.}
    \label{fig:tradeoff-stable}
    
\end{figure*}

The ratio $R_{ND} = \frac{|\text{outer steps}|}{|\text{inner steps}|}$ determines the NFEs required for each update of the Nested Diffusion intermediate prediction. Faster update rates come at the expense of lower quality in the intermediate prediction samples. To illustrate this trade-off, we present \Cref{fig:tradeoff-stable}, which showcases Nested Diffusion sampling with different $R_{ND}$ values while keeping all other hyperparameters and the random seed constant.

\begin{figure}[b]
\vskip -0.1in
\centering
    \includegraphics[width=\columnwidth]{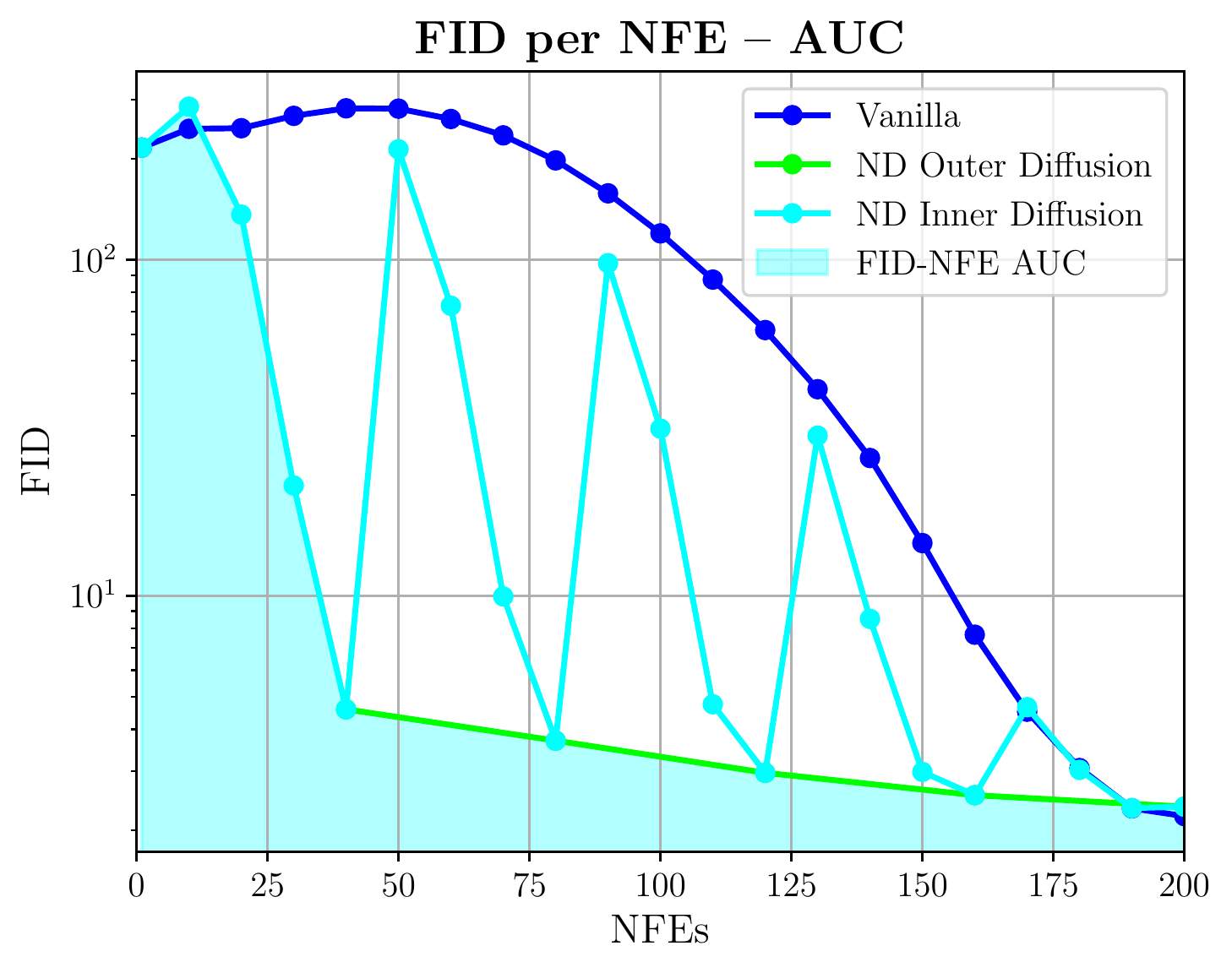}
\caption{\textbf{Graph of the AUC of the FID-NFE curve.}}
\label{fig:tradeoff-metric}

\end{figure}

To compare the performance of different Nested Diffusion hyperparameter choices, we introduce a novel metric -- the Area Under the Curve (AUC) of the log FID per NFE curve. The log FID per NFE curve is defined by the log FID of the images obtained if the algorithm were to be terminated at that particular point in the sampling process. This metric captures the intermediate FID scores, their convergence rate, as well as the frequency of the updates, thus constituting a reasonable metric for anytime generation algorithm evaluation. In the case of Nested Diffusion, the most recent $\mathcolor{purple}{\hat{\bx}_0'}$ would be returned until the termination of the first inner diffusion process. From this point, the resulting image would only be updated at the end of each subsequent  inner diffusion process. An example of this curve for Nested Diffusion, along with its corresponding AUC, is depicted in~\Cref{fig:tradeoff-metric}.

In~\Cref{tab:tradeoff-metric}, we present a comparison of various $R_{ND}$ ratios for conditional ImageNet~\cite{imagenet} generation using our proposed metric. The estimating the log FID per NFE curve is achieved by measuring 50K FID every 10 NFEs for Nested Diffusion totaling 250 NFEs. This metric captures the tradeoff between image quality and update speed, making it relevant for assessing anytime image generation algorithms. We hope this metric proves useful in comparing anytime image generation algorithms in the future. 

\begin{table}[t]
\centering
\begin{small}
\begin{sc}
\begin{tabular}{cccc}
\toprule
\textbf{Outer Steps} & \textbf{Inner Steps} & \textbf{AUC} & \textbf{Final FID}\\
\midrule
$1$	    & $200$	& $803.33$	& $2.2060$ \\
$2$	    & $100$	& $484.49$	& $\mathbf{2.1919}$ \\
$4$	    & $50$	& $354.00$	& $2.2677$ \\
$5$	    & $40$	& $\mathbf{346.15}$	& $2.3534$ \\
$10$	& $20$	& $388.18$	& $2.6267$ \\
$20$	& $10$	& $521.30$	& $3.2717$ \\
\bottomrule
\end{tabular}
\end{sc}
\end{small}
\caption{\textbf{Table of log FID per NFE AUC on ImageNet}. The result reflect different choices of inner steps and outer steps, for a total of 200 NFEs. Vanilla diffusion is equivalent to Nested Diffusion with one outer step, shown in the top line.}
\label{tab:tradeoff-metric}
\end{table}

\section{Anytime Consistency}
While maintaining consistency between intermediate samples and the final result is significant for an anytime algorithm, it's equally crucial that the anytime algorithm continues enhancing image quality during the sampling process, leading to incoherence with previous results. Based on these considerations, we conclude that it is desirable to have the semantic details in the generated image remain mostly consistent during anytime sampling, while the image itself may change. Moreover, the user should be made aware of the degree of expected change for each intermediate result produced by the algorithm, should the sampling procedure be continued.

\begin{figure*}
    \centering
    \includegraphics[width=\textwidth]{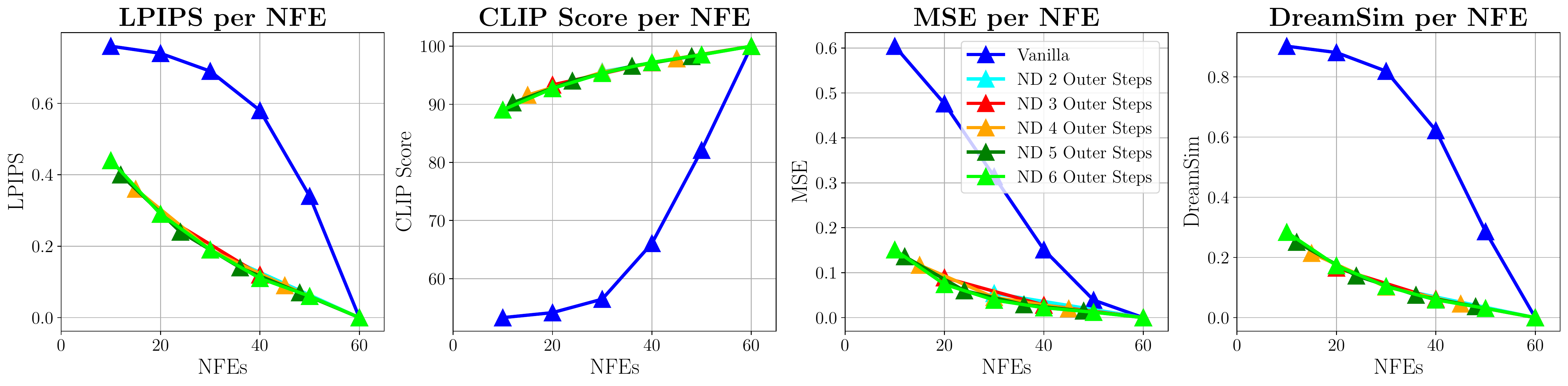}
    \caption{\textbf{Progression of distance of intermediate predictions from the final result.} Metrics are (left to right): LPIPS, image-to-image CLIP Score, MSE, and DreamSim.}
    \label{fig:constistency}
\end{figure*}

To facilitate a better understanding of the evolution in image dynamics for Nested Diffusion, the average distance of intermediate predictions from the final result is shown in~\Cref{fig:constistency}, as computed from images generated for the text-to-image experiment in~\Cref{sec:stable}. The following metrics are used; LPIPS~\cite{lpips}, image-to-image CLIP Score~\cite{hessel2021clipscore}, MSE, and DreamSim~\cite{fu2023learning}. These metrics can give an insight into the consistency dynamics, ranging from non-semantic metrics such as MSE to highly semantic metrics such as DreamSim~\cite{fu2023learning}. From the graphs, we notice that the trend is similar regardless of the choice of $R_{ND}$. The observed variance for the presented values in~\Cref{fig:constistency} is small to negligible.

\section{Implementation Details}
\subsection{Class-Conditional ImageNet Generation}
The Dit~\cite{peebles2022scalable} DNN is trained using Kullback Leibler divergence to yield both the mean and variance of a Gaussian distribution $p_\theta(\bxz|\bxt)$. The model directly predicts the conditional mean of the Gaussian noise in $\bxt$ and the variance of $p_\theta(\bxtt{t-1}|\bxt)$, but we can use a change of variables to view these as the mean and variance of $p_\theta(\bxz|\bxt)$, conforming with our notation.
When using the Dit~\cite{peebles2022scalable} DNN for Nested Diffusion, both the inner diffusion and the outer are conducted in the latent space. The variance prediction is used only in the inner diffusion, while the outer diffusion remains deterministic DDIM~\cite{song2020denoising} sampling. CFG~\cite{ho2022classifier} is regarded as part of the DNN, and therefore applied in the inner diffusion only. We set the CFG value to 1.5 similar to Peebles \& Xie, 2022~\cite{peebles2022scalable}.

\subsection{Text-to-Image Generation}
We use Stable Diffusion V1.5~\cite{latent_diffusion} FP32 to generate $512\times512$-pixel images. We implement Nested Diffusion using non-deterministic DDIM~\cite{song2020denoising} with $\eta = 0.85$ for the inner diffusion, and treat the CFG~\cite{ho2022classifier} as we did in \Cref{sec:imagenet}, setting it to the default value of 7.5. No clipping or thresholding is applied, and final $\atbar$ set to zero. Due to the size limit for submission, the images shown in the paper and supplementary material have been compressed using JPEG, which may impact the perceptual image quality. 

In the MS-COCO~\cite{lin2014microsoft} FID~\cite{fid} evaluation we follow the protocol in \cite{latent_diffusion, saharia2022photorealistic, ramesh2022hierarchical, balaji2022ediffi}, using a budget of 60 NFEs per image and using the FP16 version of Stable Diffusion V1.5 for all configurations. All other hyperparameters remain as specified above. 

The high-order solver (DPM-Solver++~\cite{lu2022dpmpp}) setup used the hyper parameters from above except for the following; The inner diffusion was based on DPM-Solver++(2S)~\cite{lu2022dpmpp}, with default hyperparamers. The outer diffusion was changed to DDIM with $\eta=\sqrtoatbar$, for larger stochasticity.

\begin{figure}[b]
\centering
    \centerline{\includegraphics[width=\columnwidth]{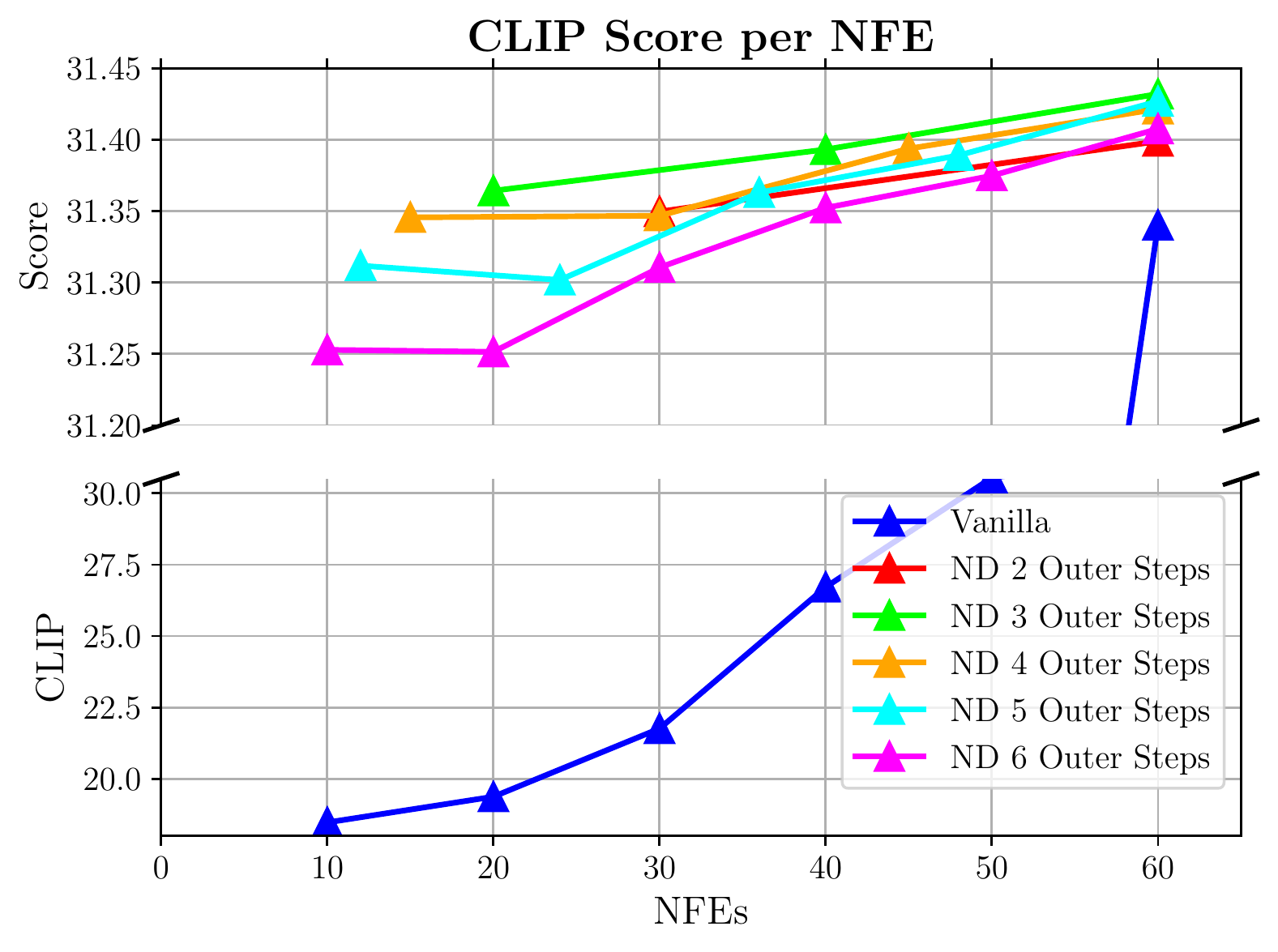}}
    \caption{\textbf{Average CLIP Scores of images generated with DDIM inner diffusion process.}}
    \label{fig:coco-clip}
    
\end{figure}

In addition to the MS-COCO FID shown in \Cref{fig:coco-fid}, we present the average CLIP Score~\cite{hessel2021clipscore} of the generated images with their guidance prompt in~\Cref{fig:coco-clip}. The CLIP Score results show a similar trend to their FID counter parts -- Nested Diffusion achieves a high score on the intermediate results and a slightly improved final image result compared to vanilla diffusion.

\begin{algorithm}
\centering
\caption{Inverse Problem Solving using Nested Diffusion}
\label{alg:inverse-nested}
\begin{algorithmic}
\State Outer diffusion denoted in \textcolor{blue}{blue}, with step size $s^o$
\State Inner diffusion denoted in \textcolor{purple}{purple}, with step sizes $\{s^i_t\}$
\State $\mathcolor{blue}{\bxtt{T}} \sim \mathcal{N}(0, \bI)$
\For{$t$ in $\{T, T-s^o,  \ldots, 1+s^o, 1\}$}
\State $\mathcolor{purple}{\bxt'} = \mathcolor{blue}{\bxt}$
\Comment{\textcolor{gray}{Beginning of inner diffusion}}
\For{$\tau$ in $\{t, t-s^i_t,  \ldots, 1+s^i_t, 1\}$}
\State $\mathcolor{purple}{\hat{\bx}_{0}'} \sim \mathcolor{purple}{p_\theta(\bxtt{0}' | \bxtt{\tau}', \by)}$
\State $\mathcolor{purple}{\bxtt{\tau - s^i_t}'} \sim \mathcolor{purple}{q'(\bxtt{\tau - s^i_t}' | \hat{\bx}_{0}', \bxtt{\tau}')}$
\EndFor
\State $\mathcolor{blue}{\bxzhat} = \mathcolor{purple}{\bxtt{0}'}$ 
\Comment{\textcolor{gray}{End of inner diffusion}}
\State $\mathcolor{blue}{\bxtt{t-s^o}} \sim \mathcolor{blue}{q(\bxtt{t-s^o} | \bxzhat, \bxt)}$
\EndFor
\State return $\mathcolor{blue}{\bxz}$
\end{algorithmic}
\end{algorithm} 

\subsection{Inverse Problem solving on CelebA-HQ256}
We evaluate the following inverse problem tasks; denoising of additive white Gaussian noise with variance set to $1.0$, block-super-resolution with factor 16, colorization, and inpainting of $50\%$ random pixels in the image. More information on these degredations can be found in DDRM~\cite{kawar2022denoising}.
our Nested Diffusion examples all use default $\eta$ hyperparamters.

The inverse problem solving algorithm using Nested Diffusion is shown in \Cref{alg:inverse-nested}. The inner diffusion is composed of a complete inverse problem sampling process (notice the similarity to \Cref{alg:nested-diffusion-sampling}). In our experiment, we have used DDRM~\cite{kawar2022denoising}, an iterative sampling process, as the aforementioned inverse problem sampling process.

\Cref{tab:inverse-table} presents PSNR and FID evaluations for Nested Diffusion on inverse problem solving. The metrics were generated on 30K samples from the CelebA-HQ256~\cite{karras2017progressive} dataset. We note that Nested Diffusion's final results are comparable to vanilla DDRM.

\begin{table}[t]
\centering
\begin{small}
\begin{sc}
\begin{tabular}{lcccc}
\toprule
\textbf{PSNR$\uparrow$} & \textbf{10 NFEs} & \textbf{20 NFEs} & \textbf{30 NFEs} & \textbf{DDRM} \\
\midrule
Denoising	&	$25.76$	&	$25.80$	&	$25.81$	&	$25.83$ \\
SR16		&	$23.19$	&	$23.55$	&	$23.89$	&	$23.78$ \\
Color		&	$19.47$	&	$21.54$	&	$21.55$	&	$23.92$ \\
Inpainting	&	$31.20$	&	$33.16$	&	$35.07$	&	$35.08$ \\
\midrule
\textbf{FID$\downarrow$} & \textbf{10 NFEs} & \textbf{20 NFEs} & \textbf{30 NFEs} & \textbf{DDRM} \\
\midrule
Denoising	&	$19.11$	&	$16.70$	&	$12.97$	&	$12.24$ \\
SR16		&	$16.77$	&	$13.79$	&	$11.65$	&	$11.30$ \\
Color		&	$12.78$	&	$7.08$	&	$6.95$	&	$4.28$ \\
Inpainting	&	$14.63$	&	$7.57$	&	$3.26$	&	$3.18$ \\
\bottomrule
\end{tabular}
\end{sc}
\end{small}
\caption{\textbf{PSNR and 30K FID of inverse problems solving on CelebA-HQ256.} The inverse problems include denoising of additive white Gaussian noise, block super-resolution with a factor of 16, colorization, and inpainting of random pixels, listed from top to bottom.}
\label{tab:inverse-table}
\end{table}

\clearpage

\onecolumn

\section{More Examples}
\label{sec:more}
We provide more examples for images generated from various experiments below.

\begin{figure*}[h]
\centering
    \includegraphics[width=6.8in]{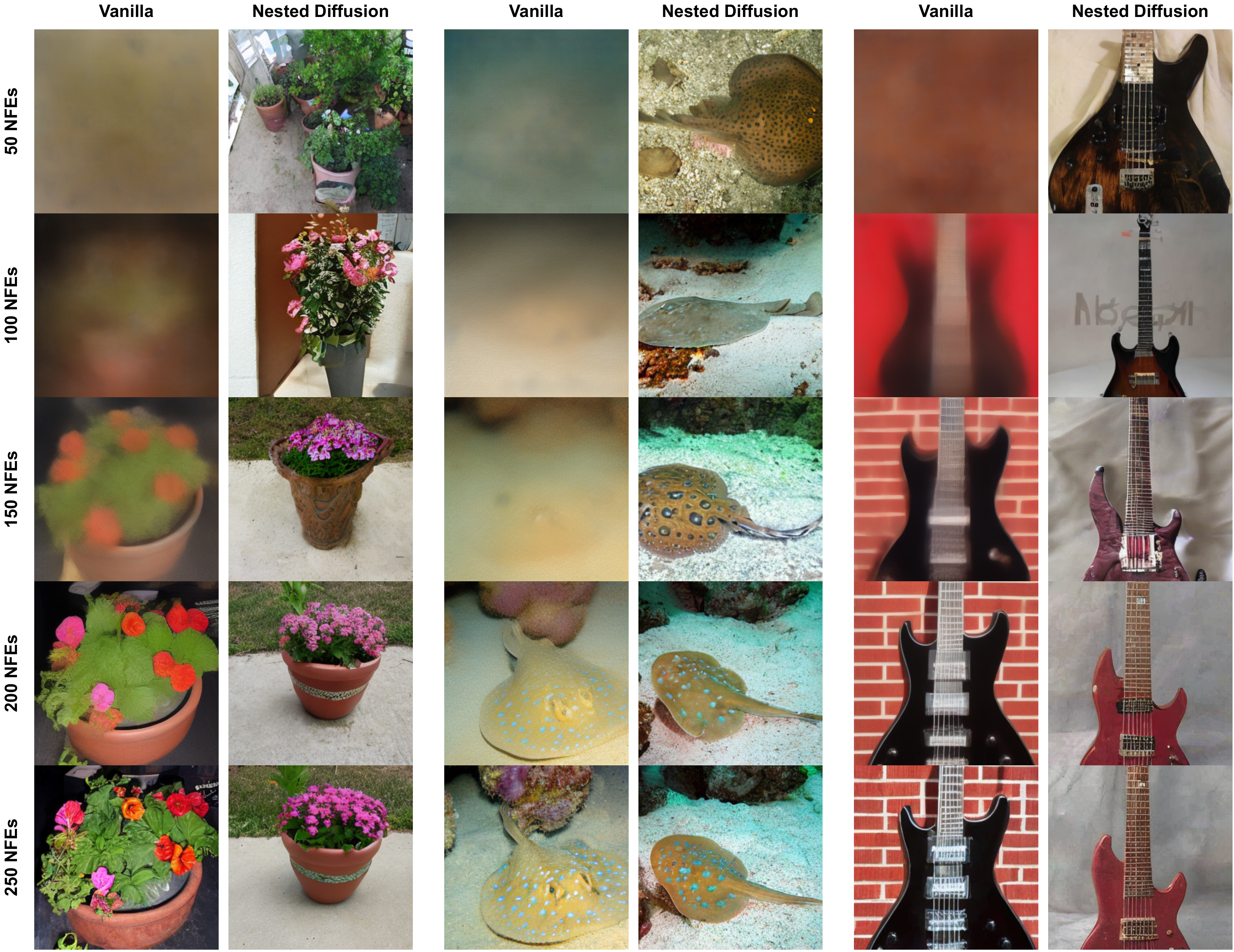}
\caption{\textbf{Additional samples of ImageNet generation, comparing vanilla diffusion model to Nested Diffusion.}}
\end{figure*}

\begin{table*}[hb]
\centering
\begin{small}
\begin{sc}
\begin{tabular}{lcccccccccccc}
\toprule
& \multicolumn{3}{c}{\textbf{Total 100 NFEs}} & \multicolumn{3}{c}{\textbf{Total 150 NFEs}} & \multicolumn{3}{c}{\textbf{Total 200 NFEs}} & \multicolumn{3}{c}{\textbf{Total 250 NFEs}} \\
\midrule
\textbf{\%} & \textbf{NFEs} & \textbf{Van} & \textbf{ND} & \textbf{NFEs} & \textbf{Van} & \textbf{ND} & \textbf{NFEs} & \textbf{Van} & \textbf{ND} & \textbf{NFEs} & \textbf{Van} & \textbf{ND}\\
\midrule
$20\%$	& $20$	& $282.89$	& $13.03$	& $30$	& $282.05$	& $6.57$ 	& $40$	& $282.83$	& $4.58$ 	& $50$	& $284.13$	& $3.57$ \\
$40\%$	& $40$	& $202.34$	& $9.20$	& $60$	& $199.74$	& $4.99$	& $80$	& $197.93$	& $3.70$	& $100$	& $197.74$	& $3.08$ \\
$60\%$	& $60$	& $65.22$	& $5.97$ 	& $90$	& $62.37$	& $3.58$ 	& $120$	& $61.82$	& $2.96$ 	& $150$	& $60.19$	& $2.61$ \\
$80\%$	& $80$	& $8.10$	& $4.00$	& $120$	& $7.67$	& $2.82$ 	& $160$	& $7.65$	& $2.54$ 	& $200$	& $7.57$	& $2.36$ \\
$100\%$	& $100$	& $2.44$	& $3.18$	& $150$	& $2.24$	& $2.50$ 	& $200$	& $2.20$	& $2.35$ 	& $250$	& $2.16$	& $2.28$ \\
\bottomrule
\end{tabular}
\end{sc}
\end{small}

\caption{\textbf{Exact 50K FID evaluation of Nested (ND) and vanilla (Van) diffusion processes}. The intermediate prediction are measure when stopped at different percentages of the full algorithm runtime (100, 150, 200, 250 NFEs).}
\label{tab:fid-imagenet-table}
\end{table*}

\begin{figure*}[t]
\centering
    \includegraphics[width=6in]{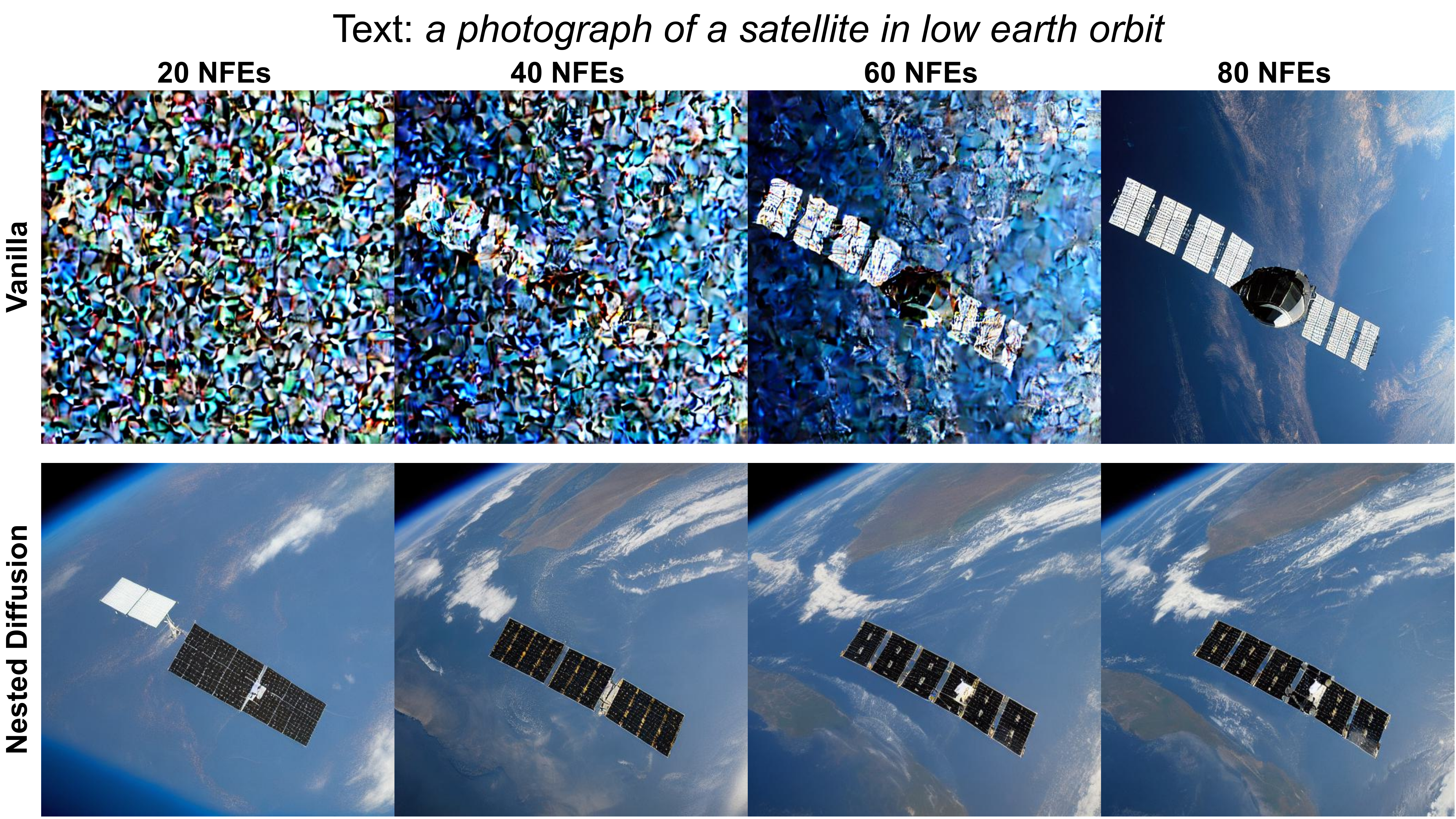}
    \vskip 0.2in
    \includegraphics[width=6in]{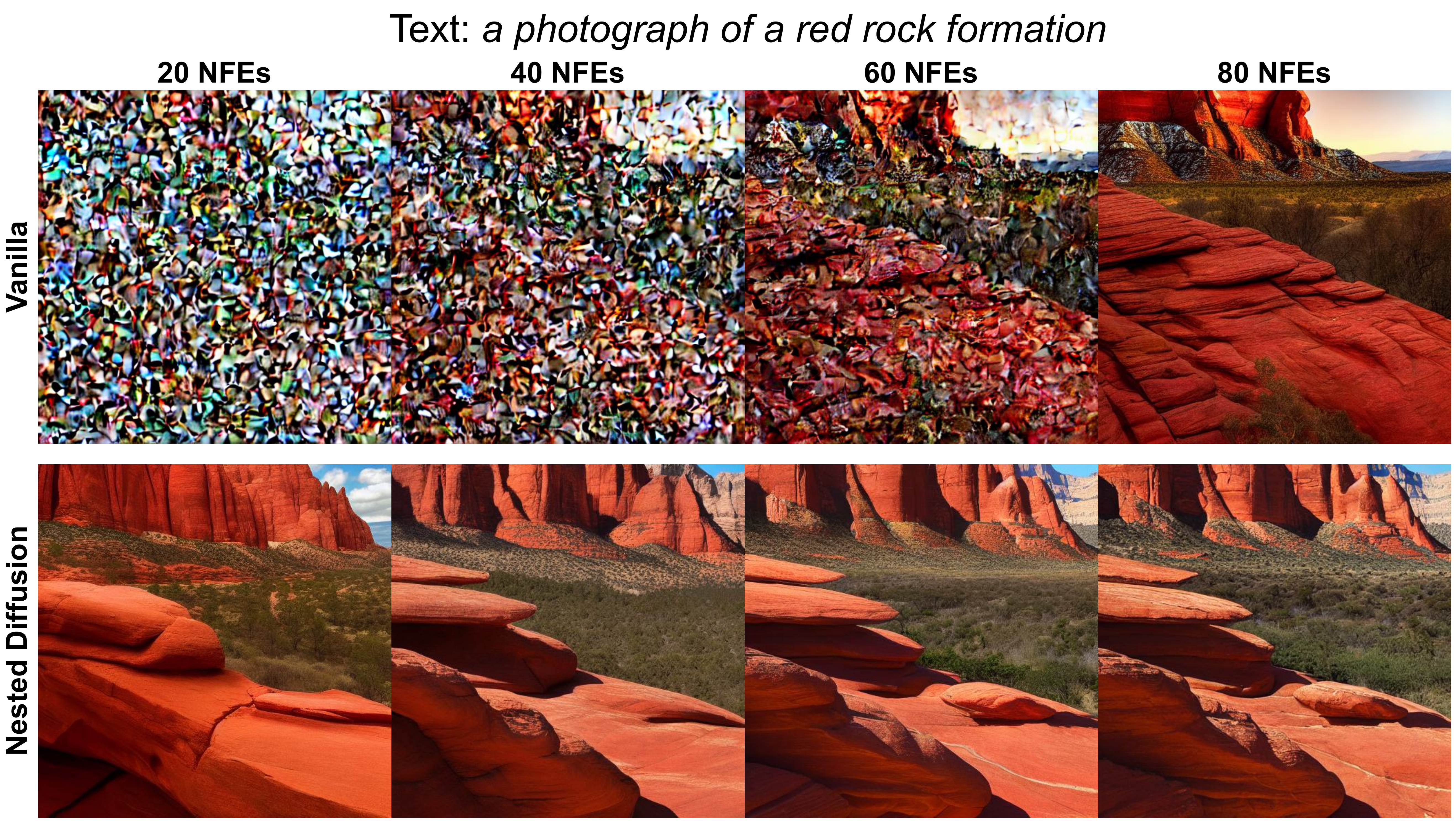}
\caption{\textbf{More Results of intermediate predictions of Stable Diffusion from a reverse diffusion process with 80 steps.}}

\end{figure*}

\begin{figure*}[t]
\centering
    \includegraphics[width=4in]{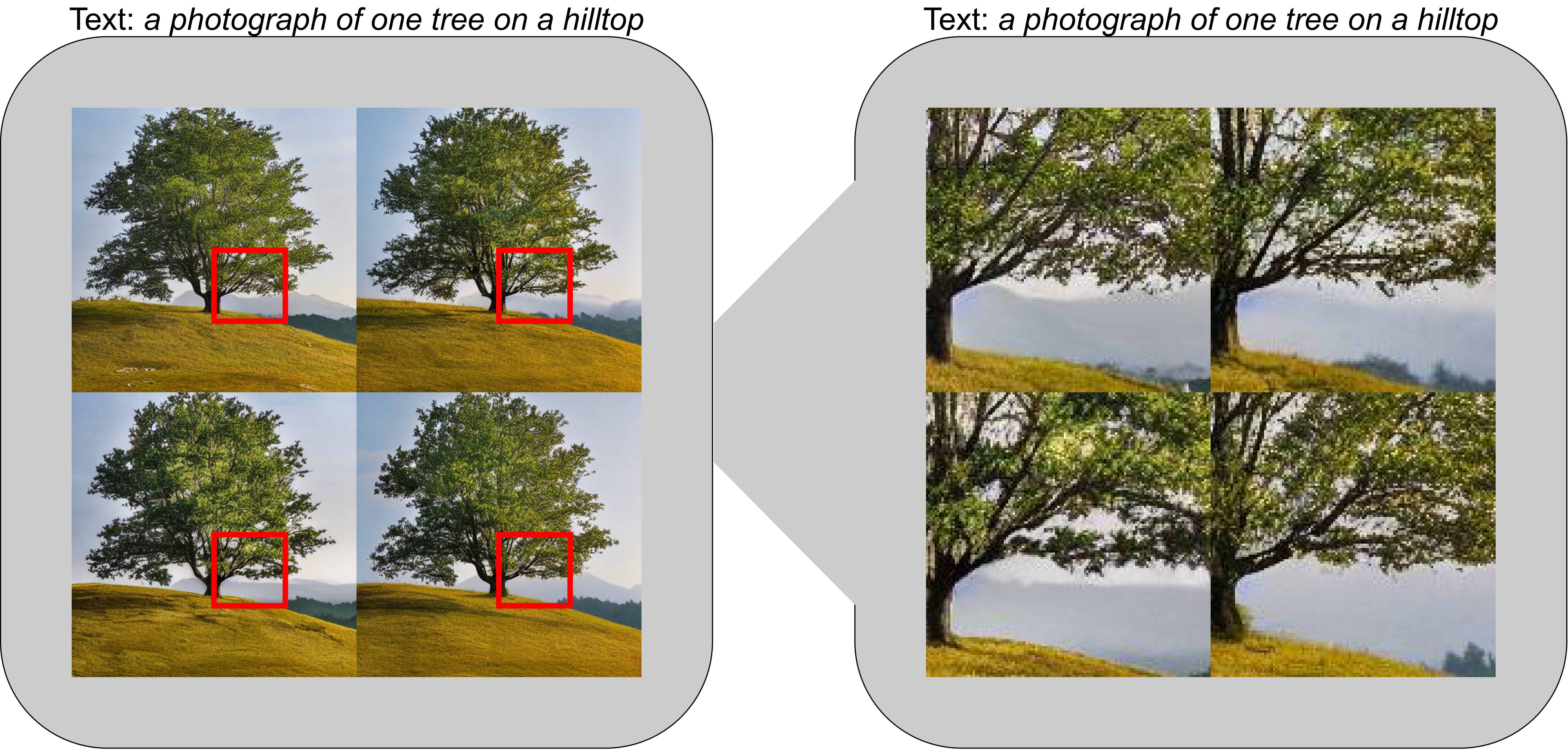}
    \vskip 0.2in
    \includegraphics[width=4in]{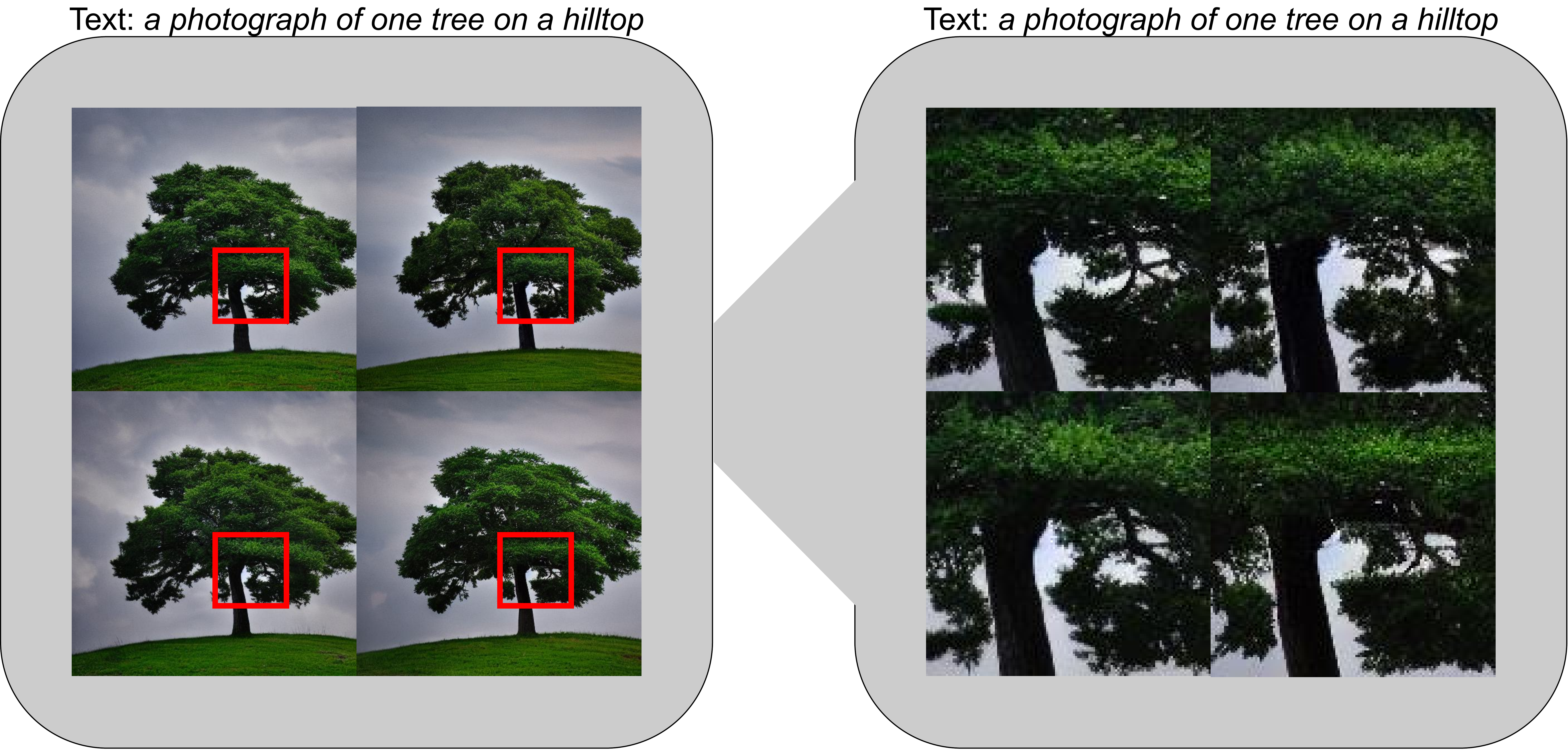}
    \vskip 0.2in
    \includegraphics[width=4in]{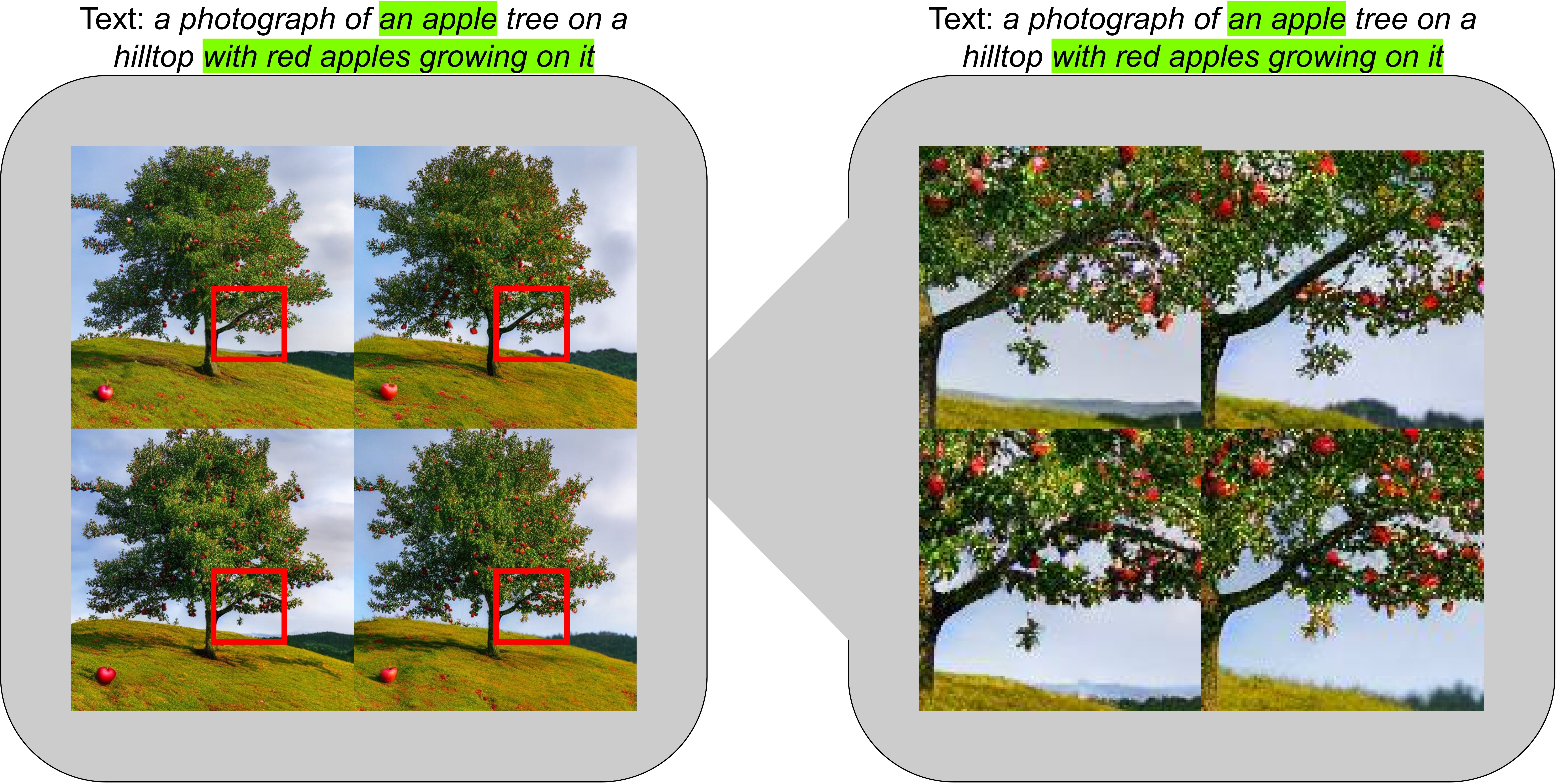}
    \vskip 0.2in
    \includegraphics[width=4in]{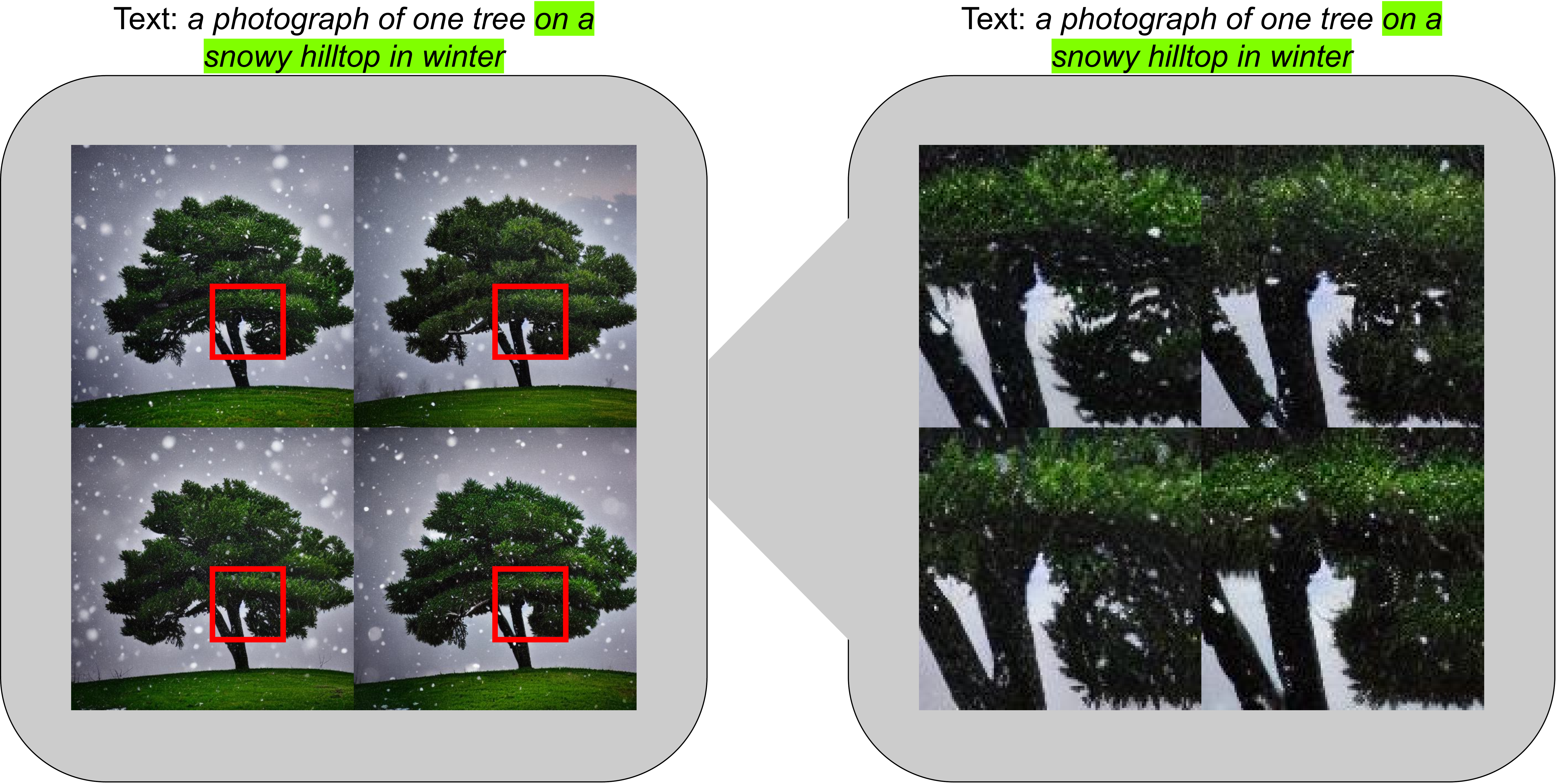}
\caption{\textbf{Zoom-in on the final images of \Cref{fig:user-guided} and \Cref{fig:user-editing} for viewing fine details in the images.}}

\end{figure*}

\end{document}